\documentclass[10pt,twocolumn,letterpaper]{article}

\usepackage{wacv}              

\usepackage{graphicx}
\usepackage{amsmath}
\usepackage{amssymb}
\usepackage{booktabs}
\usepackage{algorithm}
\usepackage{algorithmic}

\newtheorem{proposition}{Proposition}

\usepackage{booktabs}
\usepackage{multirow}
\usepackage{graphicx}
\usepackage{adjustbox}
\usepackage{amssymb}
\usepackage{pifont}
\newcommand{\cmark}{\ding{51}}%
\newcommand{\xmark}{\ding{55}}

\usepackage[pagebackref,breaklinks,colorlinks]{hyperref}

\usepackage[capitalize]{cleveref}
\crefname{section}{Sec.}{Secs.}
\Crefname{section}{Section}{Sections}
\Crefname{table}{Table}{Tables}
\crefname{table}{Tab.}{Tabs.}

\def\wacvPaperID{2028} 
\def\confName{WACV}
\def\confYear{2025}

\begin{document}

\title{F2former: When Fractional Fourier Meets Deep Wiener Deconvolution and Selective Frequency Transformer for Image Deblurring}

\author{Subhajit Paul\textsuperscript{\rm 1}, Sahil Kumawat\textsuperscript{\rm 2}, Ashutosh Gupta\textsuperscript{\rm 1}, Deepak Mishra\textsuperscript{\rm 2}\\
\textsuperscript{\rm 1}Space Applications Centre (SAC), Ahmedabad\\
\textsuperscript{\rm 2}Indian Institute of Space Science and Technology (IIST), Trivandrum\\
{\tt\small \{subhajitpaul, ashutoshg\}@sac.isro.gov.in, \{sahil.sc21b114, deepak.mishra\}@iist.ac.in}
}

\maketitle

\begin{abstract}
   Recent progress in image deblurring techniques focuses mainly on operating in both frequency and spatial domains using the Fourier transform (FT) properties. However, their performance is limited due to the dependency of FT on stationary signals and its lack of capability to extract spatial-frequency properties. In this paper, we propose a novel approach based on the Fractional Fourier Transform (FRFT), a unified spatial-frequency representation leveraging both spatial and frequency components simultaneously, making it ideal for processing non-stationary signals like images. Specifically, we introduce a Fractional Fourier Transformer (F2former), where we combine the classical fractional Fourier based Wiener deconvolution (F2WD) as well as a multi-branch encoder-decoder transformer based on a new fractional frequency aware transformer block (F2TB). We design F2TB consisting of a fractional frequency aware self-attention (F2SA) to estimate element-wise product attention based on important frequency components and a novel feed-forward network based on frequency division multiplexing (FM-FFN) to refine high and low frequency features separately for efficient latent clear image restoration. Experimental results for the cases of both motion deblurring as well as defocus deblurring show that the performance of our proposed method is superior to other state-of-the-art (SOTA) approaches.
\end{abstract}

\section{Introduction}
Image deblurring is an active research field in computer vision where it aims to restore high-quality clean data from real-world blurry observed data. It can be formulated as,
\begin{equation}
    y = k*x + n,
\end{equation}
where, $y$ is the observed blurry input image, $x$ is the clean image, $k$ refers to the blur kernel, $n$ is associated noise, and $*$ is the convolution operator. Traditional methods tackled the problem of image blur through an estimation of $k$. Among them, early blind kernel estimation methods \cite{wiener1} mainly revolve around the concept of Wiener filter \cite{wiener1949}. Later works focused on developing probabilistic approaches to formulate image priors \cite{prior1,prior2}, and optimization-oriented nonblind kernel estimation methods \cite{non-blind1,non-blind2}. Although such methods are theoretically feasible, their practical appeal is limited for real world images with large variations in blur and noise statistics.

\begin{figure}
    \centering
    \includegraphics[width=\linewidth]{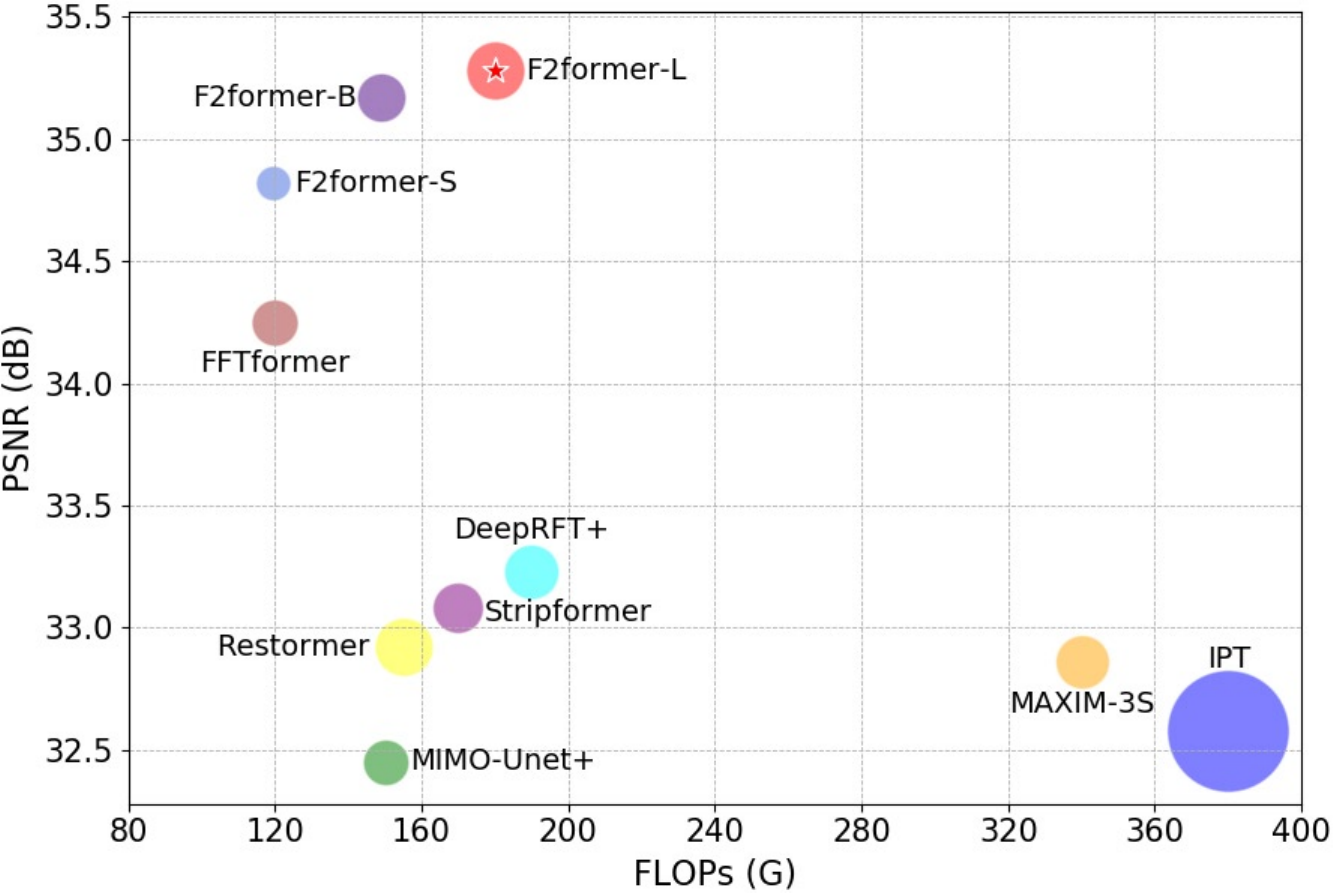}
    \caption{Comparison of the proposed model with other SOTA methods on the GoPro test dataset in terms of PSNR, floating point operations (FLOPs), and number of network parameters which are corresponded by the area of the circle.}
    \label{fig:fig1}
\end{figure}

Recent deblurring methods typically employ deep learning (DL) models that operate in both frequency and spatial domains. These methods are usually restrained from using the information of blur pattern, resulting in limited performance. Some methods \cite{ufpdeblur,wiener11} utilize this information in their DL models, achieving efficient performance. However, they are mostly effective under the uniform blur conditions. In real-world scenarios, such methods heavily fail due to the inability of the Fourier Transform (FT) to handle spatially varying non-uniform blur, as such cases require analysis of joint distribution between spatial and frequency domain information. Although short-time Fourier transform (STFT) \cite{STFT} and wavelet transform (WT) \cite{WT} serve this purpose, STFT has issues related to properly balance spatial and frequency resolution and WT suffers from optimal choice of wavelet function and higher computational complexity.

\begin{figure}
    \centering
    \includegraphics[width=\linewidth]{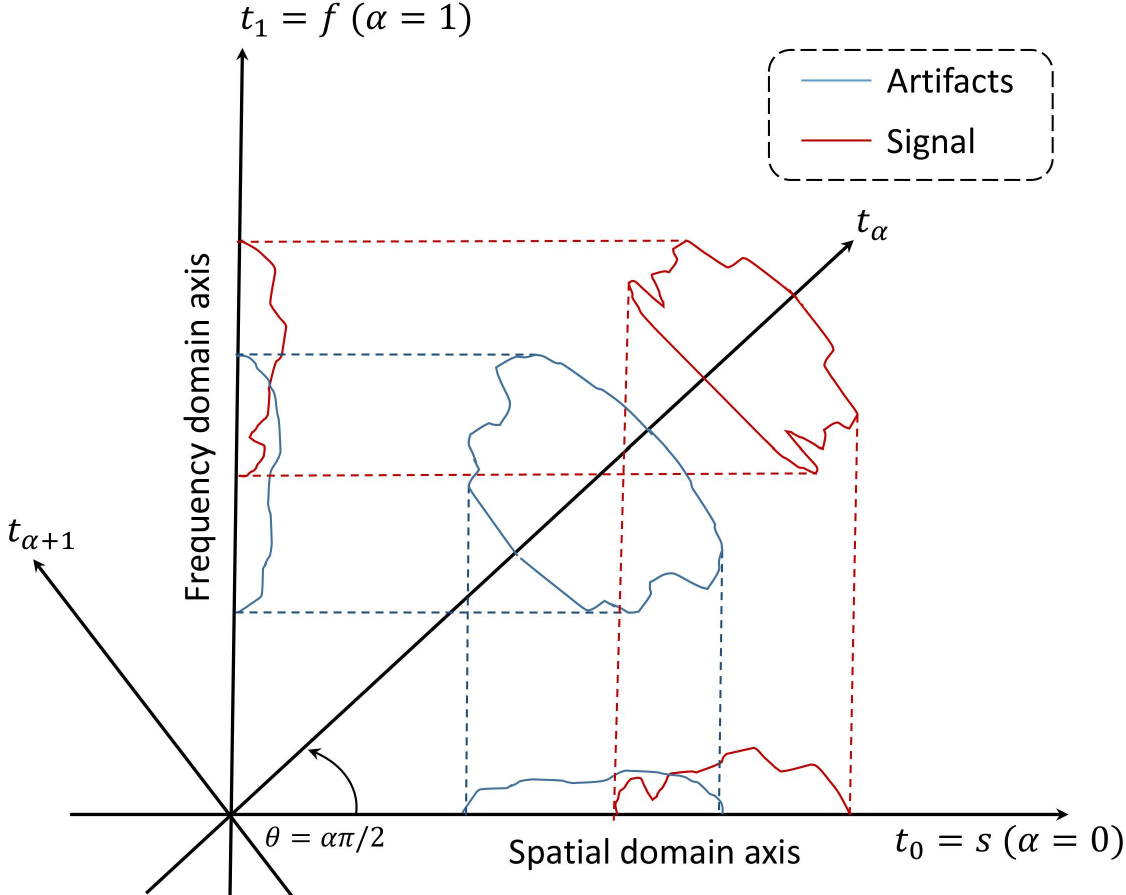}
    \caption{For optimal choice of $\alpha$, FRFT inherently separates spatially varying artefacts from non-stationary signals along the fractional order axis $t_{\alpha}$.}
    \label{fig:fig2}
\end{figure}

To address this, we explore Fractional Fourier Transform (FRFT), a generalized version of FT and a unified spatial-frequency analysis tool. This can be visualized by applying a fractional exponent $\alpha$ to the basis of regular FT providing a linear transformation with angle $\theta=\frac{\alpha\pi}{2}$ with respect to the time axis in the time-frequency domain as shown in Figure \ref{fig:fig2}. Hence, FRFT with a fractional order $\alpha$ offers an intermediate representation between spatial and frequency domains, effectively extracting spatially varying artefacts from non-stationary signals like images as shown in Figure \ref{fig:fig2}.

In this paper, we propose \textbf{F}ractional \textbf{F}ourier based trans\textbf{f}ormer (F2former) for effective image deblurring using FRFT. We design an FRFT based Wiener filter to perform deconvolution at the feature level to obtain sharp representations. To effectively reconstruct the deblurred image from these sharp features, we further employ FRFT based feature extractor and transformer blocks in a U-Net like encoder-decoder model. We develop these transformer blocks by computing frequency aware element-wise product attention at FRFT domain and extracting high and low frequency elements to selectively restore significant components in Fourier domain. Our model yields favorable results in both PSNR as well as efficiency on the benchmark dataset, GoPro as shown in Figure \ref{fig:fig1}. Our main contributions are as follows,

\begin{figure*}
    \centering
    \includegraphics[width=\textwidth]{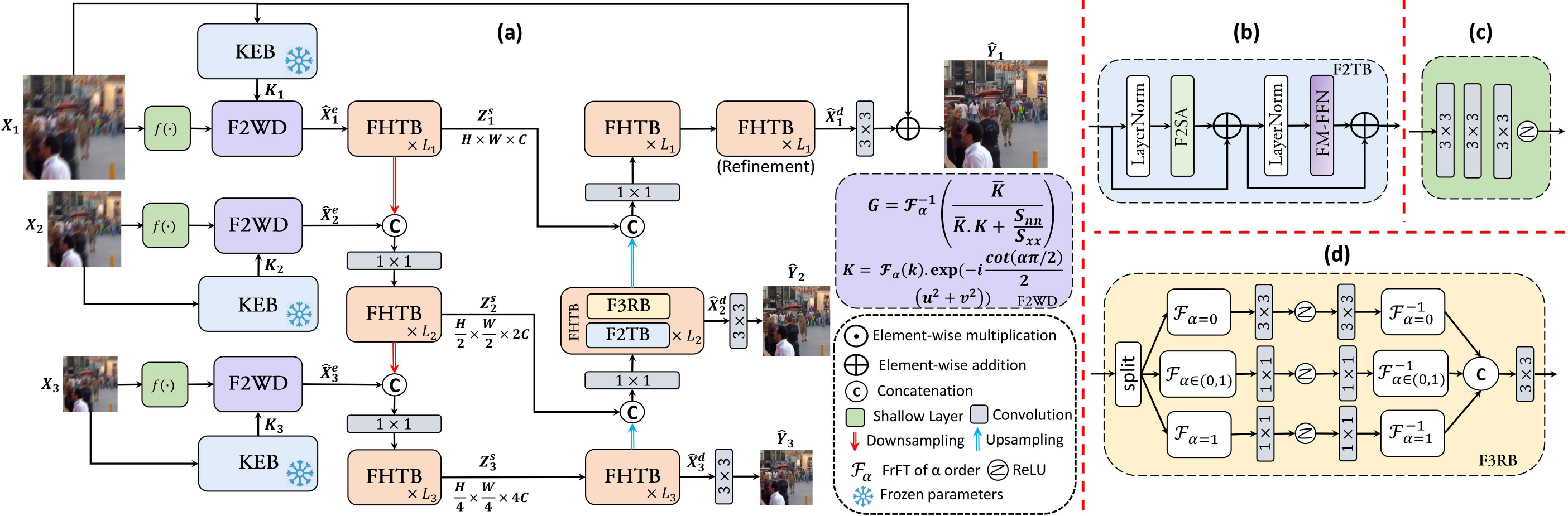}
    \caption{(a) Overall architecture of F2former. Given input $\mathbf{X}_p$ for $p$-th scale, KEB estimates the blur kernel $\mathbf{K}_p$. Given $\mathbf{X}_p$, (c) shallow layers extract features and F2WD refines them by performing deblurring using FRFT based feature Wiener deconvolution. Refined features $\mathbf{\hat{X}}_p^e$ pass through FHTB blocks for efficient image reconstruction. An FHTB consists of a FRFT based (d) feature refinement block, and $L_p$ numbers of (b) F2TBs, which consist of two major modules - self-attention estimation in FRFT domain by F2SA and frequency selective feed-forward operation by FM-FFN.}
    \label{fig:whole}
\end{figure*}

\begin{itemize}
    \item A \textit{Fractional Feature-based Wiener Deconvolution} (F2WD) layer to enhance the shallow-level features given a blurry input. As FRFT is capable of capturing spatially varying artefacts, F2WD efficiently performs deblurring in feature space.
    \item We develop a novel building block, \textit{Fractional Hybrid Transformer Block} (FHTB) to efficiently reconstruct the deblurred image. FHTB is composed of \textbf{F}ractional \textbf{F}ourier based \textbf{F}eature \textbf{R}efinement module (F3RB) and also a \textbf{T}ransformer \textbf{B}lock (F2TB) to extract local and global context features, respectively.
    \item We design F2TB with \textit{Fractional Frequency aware Self-Attention} (F2SA) for efficient computation and selective emphasis on key frequency components, and \textit{Frequency Division Multiplexing-based FFN} (FM-FFN) to dynamically extract high and low frequency features using a cosine bell function for optimal frequency recovery.
    \item We conduct experiments to demonstrate the effectiveness of the proposed F2former for motion and defocus blurring, showing significant performance improvements over other SOTA models. We also provide a detailed ablation study to validate the contribution of each module.
\end{itemize}
\section{Related Work}
\subsection{Deep learning based deblurring methods}
Recently, many  DL based effective image deblurring methods have emerged. This include both CNN based \cite{cnn1, cnn2, sfnet}, and transformer based approaches \cite{nafnet,stripformer,restormer,uformer,fftformer}. However, as convolution operation in CNN is spatially invariant, it does not capture global context for image restoration tasks resulting in limited performance. In this regard, transformer based models make substantial progress in numerous high-level vision tasks \cite{trans1,trans2,trans3} including different image restoration tasks such as deblurring, super-resolution \cite{trans4swinir}, and denoising \cite{trans5}. However, one major drawback is that transformer models suffer from the heavy computational cost of space and time complexity of $\mathcal{O}(N^2)$ and $\mathcal{O}(N^2C)$, respectively, for $N$ being number of pixels and $C$ being number of features. \cite{fftformer} brings out the concept of using FFT based element-wise product attention, producing favourable space and time complexities of $\mathcal{O}(N)$ and $\mathcal{O}(NC\log(N))$. Despite its efficacy, it really falls out in terms of feature representation, as they lacks high and low frequency feature handling. This is significant for restoration tasks as shown in \cite{sfnet,cnn3,cnn4}. The performance of these models gets further limited due to the inability of FT to handle non-stationary signals and showcase spatial-frequency properties.

\subsection{Deep Wiener deconvolution and FRFT}
Traditionally, image deblurring techniques aim to accurately estimate the blur kernel. Most DL models that learn directly with clean-degraded pairs are limited as they are unable to explicitly use this information. \cite{ufpdeblur} shows that accurate estimation of this blur pattern and incorporation of this information into latent space lead to SOTA performance in benchmark datasets. Previously, \cite{wiener11} demonstrated the effectiveness of learnable kernels in microscopy image restoration by formulating them as regularizer and integrating it in Wiener-Kolmogorov filter. Later, \cite{wiener2,wiener3} integrated classical Wiener deconvolution along with DL models. They apply Wiener deconvolution in feature space to enhance the semantic information by deblurring them in latent domain. They achieved SOTA performance when the blur kernel is known. However, these models struggle with spatially varying and non-uniform blur as they utilize only FT. Some early methods demonstrated that FRFT based image restoration techniques developed using Wiener deconvolution \cite{fractionalWiener1,fractionWiener2} can handle non-stationary signals as well as time varying characteristics much better, as the FRFT offers a more flexible and adaptive approach in analyzing and filtering such signals compared to FT. Recently, \cite{deepFrac} explored the efficacy of using FRFT in DL models achieving superior performance in different image restoration tasks. Inspired by this, in this work, we present an FRFT oriented feature-based Wiener deconvolution, along with a transformer block consisting of FRFT based self-attention and a custom FFN block to effectively work on high and low frequency features resulting in enhanced feature representation while showcasing similar complexity to \cite{fftformer}.

\section{Methodology}
\subsection{Preliminary of FRFT}
The FRFT is a generalization of the FT with a fractional parameter $\alpha \in [0,1]$. For a 1-D signal $x(s)$, where $s$ represents a spatial domain, say the FRFT of the signal is $X_{\alpha}(t_\alpha)$. Then, the $\alpha$-th order FRFT $(\mathcal{F}_{\alpha}\{\cdot\})$ and inverse FRFT (IFRFT) $(\mathcal{F}_{\alpha}^{-1}\{\cdot\})$ are defined as \cite{frft-digital},
\begin{align}\label{eq:eq1}
    &\mathcal{F}_{\alpha}\{x(s)\}(t_\alpha) = X_{\alpha}(t_\alpha) = \int_{-\infty}^{\infty} K_\alpha(s,t_{\alpha}) x(s) ds\\
    &\mathcal{F}_{\alpha}^{-1}\{X_{\alpha}(t_\alpha)\}(s) = x(s) = \int_{-\infty}^{\infty} K_{-\alpha}(s,t_{\alpha}) X_{\alpha}(t_\alpha) dt_\alpha, \notag
\end{align}
where, kernel $K_\alpha(s,t_{\alpha})$ is defined as,
\begin{equation}\label{eq:eq2}
    K_\alpha=\left\{ 
        \begin{array}{ll}
             A_{\alpha}e^{\left[\pi i \left(s^2 \cot \theta -2st_{\alpha} \csc \theta +t_{\alpha}^2 \cot \theta\right)\right]},\hspace{8mm}\theta \neq n\pi,\\
             \delta(s - t_{\alpha}),\hspace{40mm}\theta = 2n\pi, \\
             \delta(s + t_{\alpha}),\hspace{31mm}\theta = (2n+1)\pi, \\
        \end{array}
    \right.
\end{equation}
where, $A_{\alpha} = \sqrt{1-i\cot \theta}$ with $\theta = \frac{\alpha\pi}{2}$, $i=\sqrt{-1}$ and $n\in \mathbb{N}$. When $\alpha=1$, equation \ref{eq:eq1} converges to FT formulation. Another important property of FRFT is its relation with Wigner distribution, which for the signal $x(s)$ is defined as,
\begin{equation}\label{eq:eq3}
    W_x(s,f) = \int_{-\infty}^{\infty} x(s + \frac{\tau}{2})x^*(s + \frac{\tau}{2}) e^{-2\pi i f \tau}d\tau,
\end{equation}
where $f$ is frequency axis. The $W_x(s,f)$ can be interpreted as the distribution of signal energy in the spatial-frequency domain. As shown in \cite{frft-digital}, it has direct linkage with FRFT and spatial-frequency plane rotation as,
\begin{equation}\label{eq:eq4}
    W_{X_{\alpha}}(s,f) = W_x(s\cos \theta - f\sin \theta, s\sin \theta + f\cos \theta).
\end{equation}
Therefore, FRFT inherits the properties of the Wigner function resulting in high-resolution spatial-frequency representation. To apply FRFT for images, the discrete version (DFRFT) \cite{dfrft} is defined,
\begin{equation}\label{eq:eq5}
    \mathbf{F}^{\alpha}_{m.n} = \sum_{l=0, l\neq N-(N)_2}^{N} \mathbf{u}_m^l e^{(-i\frac{\alpha\pi}{2}l)}\mathbf{u}_n^l,
\end{equation}
where, $N$ is the size of the image matrix and $(N)_2$ denotes the parity, and $\mathbf{u}^l$ is $l$-th eigenvector of the normalized Discrete FT (DFT), i.e., $l$-th discrete Hermite Gaussian \cite{classFourier}. Equation \ref{eq:eq5} converges to unitary when $\alpha=0$ and to normalized DFT matrix when $\alpha=1$.
\begin{figure*}
    \centering
    \includegraphics[width=0.95\textwidth]{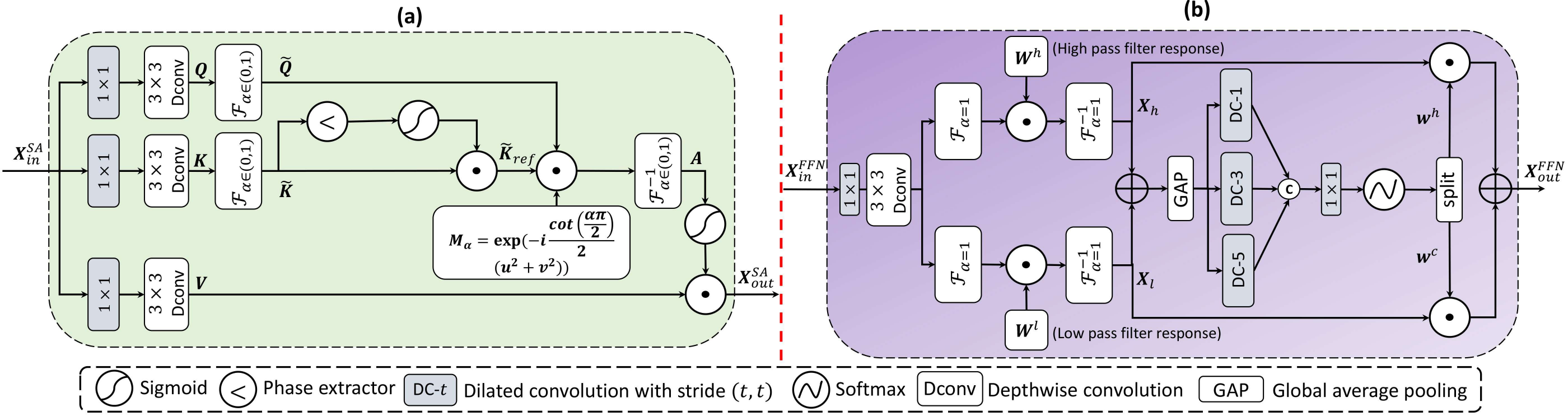}
    \caption{Architecture of F2TB layer. Overview of (a) Fractional Fourier aware Self-Attention (F2SA) and (b) Frequency division multiplexing based Feed Forward Network (FM-FFN)}
    \label{fig:transformer}
\end{figure*}
\subsection{Network architecture}
The overall framework of F2former is shown in Figure \ref{fig:whole} (a). Here, we employ a U-Net like three-scale $(p=\{1,2,3\})$ encoder-decoder architecture. Inspired by previous works \cite{cho}, we also use multi-input and multi-output strategy across different scales to overcome training difficulty. Input images of reduced sizes are fed into the model and the corresponding predicted images $(\mathbf{\hat{Y}}_p)$ are estimated by $3\times 3$ convolution after each scale in the decoder. Therefore, for a blurry image $\mathbf{X}_p \in \mathbb{R}^{\frac{H}{2^{p-1}}\times \frac{W}{2^{p-1}} \times 2^{p-1}C}$ at scale $p$, a frozen kernel estimation block (KEB) estimates its blur kernel $\mathbf{K}_p$. The KEB is modelled based on \cite{ufpdeblur} and \cite{keb}. More details about KEB are provided in implementation details in Appendix A.3 of supplementary. Given estimated $\mathbf{K}_p$, and $f(\mathbf{X}_p)$, F2WD performs deblurring at feature level to enhance the low-level feature representations, where $f(\cdot)$ is shallow layer consisting of CNN layers as shown in Figure \ref{fig:whole} (c). The refined features $\mathbf{\hat{X}}_p^e$ are passed to the main encoder-decoder model. At each scale of this model, there is an FHTB block consisting of an F3RB layer followed by $L_p$ number of F2TB layers. F3RB is designed inspired by \cite{deepFrac} to extract features at different context level by implying FRFT followed by CNN for different $\alpha$ values as shown in Figure \ref{fig:whole} (d). However, here we also employ ReLU activations before IFRFT as they refine features by selecting certain frequency components \cite{intriguing}. We also use skip connections between encoder and decoder features to assist in effective gradient flow during training, and implement upsampling and downsampling via pixel shuffle and unshuffle operations. Next, we discuss the functionality of F2WD and F2TB in detail.
\subsubsection{Fractional Feature based Wiener Deconvolution (F2WD)}
\cite{wiener2} shows for a set of linear filters, the derivation of the feature-based Wiener deconvolution (FWD) operator is similar to the classical one. Although deep CNN features are non-linear, they are locally linear \cite{random1}. Therefore, \cite{wiener2} estimates the blurry features using deep convolution layers and performs Wiener deconvolution in the feature space. They further demonstrate the advantage of performing the FWD operation in deep feature space instead of image space. Inspired by this, we introduce F2WD to refine the features at the shallow level.

The major issue in  \cite{wiener2} is limited performance for non-uniform blur, as for arbitrary spatially varying degradation or nonstationary processes like images, the derived deconvolution operator cannot be realized adequately in Fourier domain due to their time-variant properties. As these properties are more realizable in fractional Fourier domain, we derive the FWD operator $(\mathbf{G})$ using FRFT \cite{fractionalWiener1} as shown in proposition 1.
\begin{proposition}
    \textit{Consider $\mathbf{K}_p$ be the estimated blur kernel. Then, for the $j$-th feature, the FWD operator in the FRFT domain can be estimated as}
    \begin{equation}\label{eq:eq6}
        \mathbf{G}_j^p = \frac{(\mathbf{C}\mathbf{F}^{\alpha} \mathbf{S}_j^{xx} \mathbf{K}_p^H \mathbf{F}^{-\alpha})^H}{\mathbf{F}^{\alpha}(\mathbf{K}_p\mathbf{S}_j^{xx}\mathbf{K}_p^H +\mathbf{S}_j^{nn})\mathbf{F}^{-\alpha}},
    \end{equation}
    \textit{where, $(\cdot)^H$ is conjugate transpose operation, $\mathbf{S}_j^{xx}$ and $\mathbf{S}_j^{nn}$ is auto-correlation of clean signal and noise content for $j$-th feature, $\mathbf{C}$ is a diagonal matrix corresponds to chirp multiplication with diagonal elements} $\mathbf{C}_{kk} = e^{-ik^2\cot \theta/2}$.
\end{proposition}
Detailed proof of the above proposition is provided in the Appendix B of supplementary. Here, we follow \cite{wiener2} to estimate $\mathbf{S}_j^{xx}$ and $\mathbf{S}_j^{nn}$. Hence, for each scale $p$, F2WD performs deconvolution using the above derived operator $(\mathbf{G}^p)$ with low level features $f(\mathbf{X}_p)$ such that $\mathbf{\hat{X}}_p^e = \mathbf{G}^p f(\mathbf{X}_p)$as shown in Figure \ref{fig:whole} (a). To show the efficacy of the proposed F2WD, we have presented a comparative feature visualization in the ablation study.

\subsubsection{Fractional frequency based transformer block (F2TB)}
Each F2TB consists of two important modules, F2SA and FM-FFN as shown in Figure \ref{fig:whole} (a).\par
\textbf{Fractional Frequency aware Self-Attention (F2SA).} The detail architectural of F2SA are shown in Figure \ref{fig:transformer} (a).  The major issue with transformers is their computational overhead in self-attention (SA) layer, as time and space complexities increase quadratically $(\mathcal{O}(N^2))$. \cite{fftformer} introduces query-key Hadamard product attention in Fourier domain as an alternate resulting in a version with $\mathcal{O}(N\log N)$ complexity. Here, we present a similar attention mechanism, but in the FRFT domain with the awareness of important spatial-frequency components.

For an input $\mathbf{X}_{in}^{SA}\in \mathbb{R}^{H\times W \times C}$, F2TB estimates query $(\mathbf{Q})$, key $(\mathbf{K})$ and value $(\mathbf{V})$ matrix by linear projections, \textit{i.e.}, $\mathbf{Q} = W_q\mathbf{X}_{in}^{SA}$, $\mathbf{K} = W_k\mathbf{X}_{in}^{SA}$, and $\mathbf{V} = W_v\mathbf{X}_{in}^{SA}$, where, $W_{(\cdot)}$ corresponds to $1\times 1$ convolution followed by $3\times 3$ depth-wise convolution. During estimating Hadamard-product attention, we follow the convolution theorem associated with FRFT \cite{randomFrft} as follows
\begin{equation}\label{eq:eq7}
    \mathcal{F}_{\alpha} \{x(s) * y(s)\}(t_{\alpha}) = e^{-it_{\alpha}^2\cot \theta/2}X(t_{\alpha})Y(t_{\alpha})
\end{equation}

Hence, during element-wise attention estimation, we also have to consider the chirp equivalent matrix $\mathbf{M}_{\alpha} = \mathbf{m}_H\mathbf{m}_W^T$, where $\mathbf{m}_N = [e^{-i.1^2.\cot \theta/2} \quad e^{-i.2^2.\cot \theta/2} \quad \ldots \quad e^{-i.N^2.\cot \theta/2}]^T$.\par
\begin{table*}
\centering
\caption{Quantitative comparison of SOTA methods on GoPro, HIDE, RealBlur-R, and RealBlur-J datasets for setting $\mathcal{A}$. The first and second best results are in bold and underlined. Inference time is based on testing of $256\times 256$ image patch. }
\begin{adjustbox}{max width=0.86\textwidth}
\begin{tabular}{lcccccccccc}
\toprule
\multirow{2}{*}{Method} & \multicolumn{2}{c}{GoPro} & \multicolumn{2}{c}{HIDE} & \multicolumn{2}{c}{RealBlur-R} & \multicolumn{2}{c}{RealBlur-J} & \multirow{2}{*}{\begin{tabular}[c]{@{}c@{}}Avg. \\ Runtime (s) \end{tabular}} & \multirow{2}{*}{\begin{tabular}[c]{@{}c@{}}Parameters \\ (M) \end{tabular}} \\
\cmidrule(r){2-3} \cmidrule(r){4-5} \cmidrule(r){6-7} \cmidrule(r){8-9}
 & PSNR $\uparrow$ & SSIM $\uparrow$ & PSNR $\uparrow$ & SSIM $\uparrow$ & PSNR $\uparrow$ & SSIM $\uparrow$ & PSNR $\uparrow$ & SSIM $\uparrow$ & & \\
\midrule
DMPHN \cite{dbgan}& 31.20 & 0.940 & 29.09 & 0.924 & 35.70 & 0.948 & 28.42 & 0.860 & 0.21 & 21.7 \\
DBGAN \cite{dbganv2}& 31.10 & 0.942 & 28.94 & 0.915 & 35.78 & 0.909 & 24.93 & 0.745 & 0.04 & 60.9 \\
MT-RNN \cite{mt-rnn} & 31.15 & 0.945 & 29.15 & 0.918 & 35.79 & 0.951 & 28.44 & 0.862 & - & - \\
MPRNet \cite{cnn2} & 32.66 & 0.959 & 30.96 & 0.939 & 35.99 & 0.952 & 28.70 & 0.873 & 0.09 & 20.1  \\
MIMO-UNet+ \cite{mimo}& 32.45 & 0.957 & 29.99 & 0.930 & 35.54 & 0.947 & 27.63 & 0.837 & - & 16.1 \\
Uformer \cite{uformer} & 33.06 & 0.967 & 30.90 & 0.953 & 36.19 & 0.956 & 29.09 & 0.886 & 0.07 & 50.9 \\
NAFNet64 \cite{nafnet} & 33.69 & 0.967 & 31.32 & 0.943 & 35.84 & 0.952 & 29.94 & 0.854 & 0.04 & 67.9 \\
Stripformer \cite{stripformer}& 33.30 & 0.966 & 31.16 & 0.950 & - & - & - & - & 0.04 & 19.7 \\
Restormer \cite{restormer} & 32.92 & 0.961 & 31.22 & 0.942 & 36.19 & 0.957 & 28.96 & 0.879 & 0.08 & 26.1 \\
DeepRFT+ \cite{intriguing} & 33.52 & 0.965 & 31.66 & 0.946 & 36.11 & 0.955 & 28.90 & 0.881 & 0.09 & 23.0 \\
FFTformer \cite{fftformer} & 34.21 & 0.968 & 31.62 & 0.946 & - & - & - & - & 0.13 & 16.6 \\
UFPNet \cite{ufpdeblur} & 34.06 & 0.968 & 31.74 & 0.947 & 36.25 & 0.953 & 29.87 & 0.884 & - & 80.3 \\
MRLPFNet \cite{lpfnet}& 34.01 & 0.968 & 31.63 & 0.947 & - & - & - & - & 0.863 & 20.6 \\
AdaRevD \cite{xintm2024AdaRevD}& \underline{34.64} & \underline{0.972} & \underline{32.37} & \underline{0.953} & \underline{36.60} & \underline{0.958} & \underline{30.14} & \underline{0.895} & - & - \\
\midrule

F2former-S (ours) & \textbf{34.77} & 0.963 & \textbf{33.07} & \textbf{0.958} & \textbf{36.64} & \textbf{0.959} & \textbf{30.88} & \textbf{0.907} & 0.15 & 9.29 \\

F2former-B (ours) & \textbf{35.17} & 0.970 & \textbf{33.43} & \textbf{0.966} & \textbf{36.84} & \textbf{0.961} & \textbf{31.25} & \textbf{0.918} & 0.22 & 18.3 \\

F2former-L (ours) & \textbf{35.22} & \textbf{0.975} & \textbf{33.48} & \textbf{0.971} & \textbf{36.88} & \textbf{0.967} & \textbf{31.34} & \textbf{0.920} & 0.27 & 26.2 \\
\bottomrule
\end{tabular}
\end{adjustbox}
\label{tab:go-pro}
\end{table*}

\begin{figure*}
    \centering
    \includegraphics[width=\textwidth]{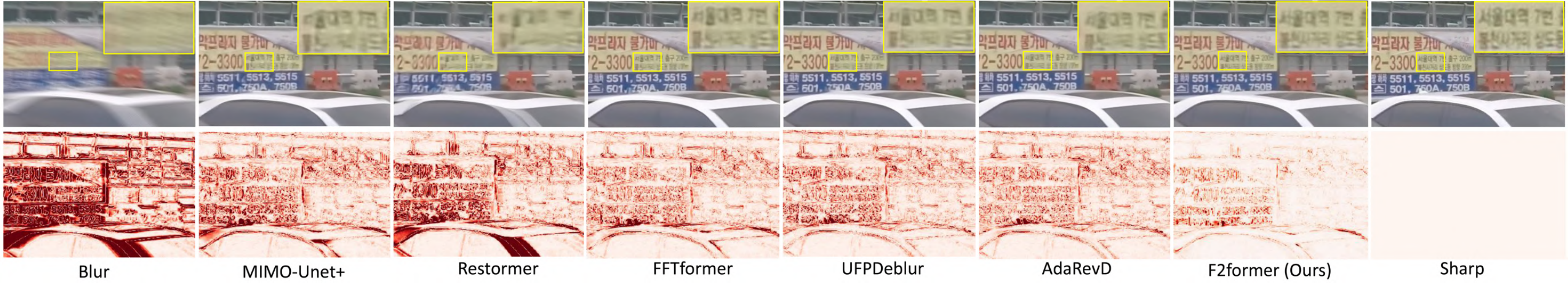}
    \caption{Visual illustration on the GoPro test dataset. The first row shows blurred image, predicted images of various methods, and GT sharp image. The second row shows the residual of the respective blurred image and predicted sharp images with GT.}
    \label{result:go_pro}
\end{figure*}
For attention estimation, the projected $\mathbf{Q}$ and $\mathbf{K}$ in FRFT domain can be written as $\mathbf{\tilde{Q}} = \mathbf{F}^{\alpha}\mathbf{Q}$ and $\mathbf{\tilde{K}} = \mathbf{F}^{\alpha}\mathbf{K}$. Instead of directly using $\mathbf{\tilde{K}}$, we refine it based on important frequency components. As the phase of FRFT consists of relevant spatial-frequency information \cite{frftphase}, we use that as a modulator to update $\mathbf{\tilde{K}}$. As illustrated in Figure \ref{fig:transformer} (a), the refined key in the FRFT domain is estimated as 
\begin{equation}\label{eq:eq8}
    \mathbf{\tilde{K}}_{ref} = \sigma(\arg (\mathbf{\tilde{K}})) \odot \mathbf{\tilde{K}}, 
\end{equation}
where $\arg(\cdot)$ represents phase component, $\odot$ is Hadamard product, and $\sigma(\cdot)$ is Sigmoid operation. Then, the attention can be estimated as 
\begin{equation}\label{eq:eq9}
    \mathbf{A} = \mathbf{F}^{-\alpha} (\mathbf{M}_{\alpha} \odot (\mathbf{\tilde{Q}} \odot \mathbf{\tilde{K}}_{ref}^H)^H).
\end{equation}
Finally, the output of F2SA is estimated as
\begin{equation}\label{eq:eq10}
    \mathbf{X}_{out}^{SA} = \mathbf{X}_{in}^{SA} + conv(\sigma(A) \odot \mathbf{V}),
\end{equation}
where, $conv(\cdot)$ is $1\times 1$ CNN operation. For computing attention, we divide the number of channels into heads to learn them parallelly similar to conventional multi-head SA \cite{randomTrans}. For the optimal choice of $\alpha$ in F3RB, F2WD and F2SA, we use it as a learnable parameter as suggested by \cite{trainableFRFT}.

\textbf{Frequency division Multiplexing based FFN (FM-FFN).} We design FFN to further refine the features from F2SA based on selective frequency components as shown in Figure \ref{fig:transformer} (b). Given input $\mathbf{X}_{in}^{FFN}\in \mathbb{R}^{H\times W \times C}$, we re-project them using $W_{FFN}$ and apply high and low pass filter operation in FFT domain to divide the features into different frequency components. Hence, the estimated high $(\mathbf{X}_h)$ and low $(\mathbf{X}_l)$ frequency components can be expressed as
\begin{equation}\label{eq:eq11}
    \begin{aligned}
        &\mathbf{X}_h = \mathbf{F}^{\alpha} (\mathbf{W}^h \odot (W_{FFN}\mathbf{\mathbf{X}_{in}^{FFN}}))\mathbf{F}^{-\alpha},\\
        &\mathbf{X}_l = \mathbf{F}^{\alpha} (\mathbf{W}^l \odot (W_{FFN}\mathbf{\mathbf{X}_{in}^{FFN}}))\mathbf{F}^{-\alpha},
    \end{aligned}
\end{equation}
where, we set $\alpha=1$ for operation in FT domain, and $\mathbf{W}^h$ and $\mathbf{W}^l$ correspond to high and low pass filter response. Here, we prefer FT as it is easy to separate different frequency components in a frequency only representation. The $\mathbf{W}^l$ is estimated using the cosine bell function (CBF) defined as follows,
\begin{equation}\label{eq:eq12}
    \mathbf{W}_{j,k}^l = 0.5\left(1+ \cos \left( \frac{\pi\sqrt{j^2 + k^2}}{\lambda}\right) \right),
\end{equation}
where $\lambda$ is learnable low-pass cut-off frequency. From this, the high pass response is estimated as $\mathbf{W}^h = \mathbf{I} - \mathbf{W}^l$ with $\mathbf{I}$ being the identity matrix. The main advantage of CBF is its smooth transition, resulting in fewer ringing artefacts during inverse operation. We validate this with our ablation study. 

To emphasize important frequency components, we leverage frequency multiplexing operation based on \cite{randKernel}. Specifically, to get channel-wise weights related to significant frequency elements, we perform $\mathbf{Z} = GAP(\mathbf{X}_h + \mathbf{X}_l)$ such that $\mathbf{Z}\in \mathbb{R}^{1\times1\times C}$, where $GAP(\cdot)$ is global average pooling, followed by $3\times 3$ CNN with different dilation rates of 3, 5, and 7 as shown in Figure \ref{fig:transformer} (b) to extract features at different receptive fields. They are later concatenated and passed to $1\times 1$ CNN and Softmax activation, formulated as, $(\mathbf{w}^h \oplus \mathbf{w}^l)_c = \frac{e^{(W_D\mathbf{Z})_c}}{\sum_j^{2C}e^{(W_D\mathbf{Z})_j}}$, where $W_D$ corresponds to all intermediate CNN operation, $\mathbf{w}^h$ and $\mathbf{w}^l$ are channel-wise attention weights for low and high frequency components, and $\oplus$ represents concatenation. The final weights can be obtained by split operation. Therefore, the final output of FM-FFN can be written as,
\begin{equation}\label{eq:eq13}
    \mathbf{X}_{out}^{FFN} = \mathbf{X}_{in}^{FFN} + conv((\mathbf{w}^l \odot \mathbf{X}_l) + (\mathbf{w}^h \odot \mathbf{X}_h)).
\end{equation}

Due to page limitations, we discuss the objective functions and implementation details to train our network in Appendix A.1 of supplementary.

\section{Experimental Results}
\subsection{Dataset and Experimental Setup}
We investigate the effectiveness of the proposed method for both motion blur and spatially varying non-uniform blur by evaluating five different commonly used benchmark datasets: GoPro \cite{GoPro}, HIDE \cite{HIDE}, RealBlur-R and RealBlur-J \cite{realBlur}, and Dual-Pixel Defocus Deblurring (DPDD) \cite{DDPD}. For detailed analysis, we introduce three F2former variants, F2former-S (Small), F2former-B (Base), and F2former-L (Large) based on feature channels $C$ and number of Transformer blocks in FHTB, $L=\{L_1, L_2, L_3\}$ as shown in Figure \ref{fig:whole} (a). For all three variants, we take $C=48$. For base model, we select $L=\{2,4,8\}$, while for large and small model we take $L=\{4,8,12\}$ and $L=\{2,2,2\}$, respectively. The details about other parameter settings are given in the Appendix A.2 of supplementary. 

\begin{table*}
    \centering
    \caption{Defocus deblurring comparisons on the DPDD testset (containing 37 indoor and 39 outdoor scenes) for setting $\mathcal{B}$. \textbf{S}: single-image defocus deblurring. \textbf{D}: dual-pixel defocus deblurring.}  
    \small
    \begin{adjustbox}{max width=0.86\textwidth}
    \begin{tabular}{lcccccccccccc}
        \toprule
        \multirow{2}{*}{Method} & \multicolumn{4}{c}{Indoor Scenes} & \multicolumn{4}{c}{Outdoor Scenes} & \multicolumn{4}{c}{Combined} \\
        \cmidrule(r){2-5} \cmidrule(r){6-9} \cmidrule(r){10-13}
        & PSNR $\uparrow$ & SSIM $\uparrow$ & MAE $\downarrow$ & LPIPS $\downarrow$ & PSNR $\uparrow$ & SSIM $\uparrow$ & MAE $\downarrow$ & LPIPS $\downarrow$ & PSNR $\uparrow$ & SSIM $\uparrow$ & MAE $\downarrow$ & LPIPS $\downarrow$ \\
        \midrule
        DMENet$_S$ \cite{Lee2019DMENet} & 25.50 & 0.788 & 0.038 & 0.298 & 21.43 & 0.644 & 0.063 & 0.397 & 23.41 & 0.714 & 0.051 & 0.349 \\
        KPAC$_S$ \cite{kpac}& 27.97 & 0.852 & 0.026 & 0.182 & 22.62 & 0.701 & 0.053 & 0.269 & 25.22 & 0.774 & 0.040 & 0.227 \\
        IFAN$_S$ \cite{ifan}& 28.11 & 0.861 & 0.026 & 0.179 & 22.76 & 0.720 & 0.052 & 0.254 & 25.37 & 0.789 & 0.039 & 0.217 \\
        SFNet$_S$\cite{sfnet} & 29.16 & 0.878 & 0.023 & 0.168 & 23.45 & 0.747 & 0.049 & 0.244 & 26.33 & 0.811 & 0.037 & 0.207 \\
        Restormer$_S$ \cite{restormer}& 28.87 & 0.882 & 0.025 & 0.145 & 23.24 & 0.743 & 0.050 & 0.209 & 25.98 & 0.811 & 0.038 & 0.178 \\
        GRL-B$_S$ \cite{grl}& \underline{29.06} & \underline{0.886} & \underline{0.024} & \underline{0.139} & \underline{23.45} & \underline{0.761} & \underline{0.049} & \underline{0.196} & \underline{26.18} & \underline{0.822} & \underline{0.037} & \underline{0.168} \\
        F2former$_S$ (ours) & \textbf{31.18} & \textbf{0.905} & \textbf{0.011} & \textbf{0.109} & \textbf{26.60} & \textbf{0.814} & \textbf{0.032} & \textbf{0.134} & \textbf{28.89} & \textbf{0.865} & \textbf{0.022} & \textbf{0.121} \\
        \midrule
        RDPD$_D$ \cite{rdpd} & 28.10 & 0.843 & 0.027 & 0.210 & 22.82 & 0.704 & 0.053 & 0.298 & 25.39 & 0.772 & 0.040 & 0.255 \\
        Uformer$_D$\cite{uformer} & 28.23 & 0.859 & 0.026 & 0.199 & 22.95 & 0.728 & 0.051 & 0.285 & 25.59 & 0.795 & 0.039 & 0.243 \\
        IFAN$_D$ \cite{ifan} & 28.66 & 0.870 & 0.025 & 0.172 & 23.46 & 0.743 & 0.049 & 0.254 & 25.99 & 0.804 & 0.037 & 0.207 \\
        Restormer$_D$\cite{restormer} & 29.48 & 0.895 & 0.023 & 0.134 & 23.97 & 0.773 & 0.047 & 0.175 & 26.66 & 0.833 & 0.035 & 0.155 \\
        LAKDNet$_D$ \cite{lakdnet} & 29.76 & 0.893 & 0.023 & \underline{0.109} & \underline{24.42} & 0.765 & \underline{0.045} & \underline{0.142} & 27.02 & 0.829 & 0.034 & 0.140 \\
        GRL-B$_D$ \cite{grl} & \underline{29.83} & \underline{0.903} & \underline{0.022} & 0.114 & 24.39 & \underline{0.795} & \underline{0.045} & 0.150 & \underline{27.04} & \underline{0.847} & \underline{0.034} & \underline{0.133} \\
        F2former$_D$ (ours) & \textbf{32.06} & \textbf{0.920} & \textbf{0.015} & \textbf{0.097} & \textbf{27.19} & \textbf{0.825} & \textbf{0.028} & \textbf{0.119} & \textbf{29.63} & \textbf{0.875} & \textbf{0.022} & \textbf{0.108} \\
        \bottomrule 
    \end{tabular}
    \end{adjustbox} 
    \label{tab:ddpd}
\end{table*}
\begin{figure*}
    \centering
    \includegraphics[width=\textwidth]{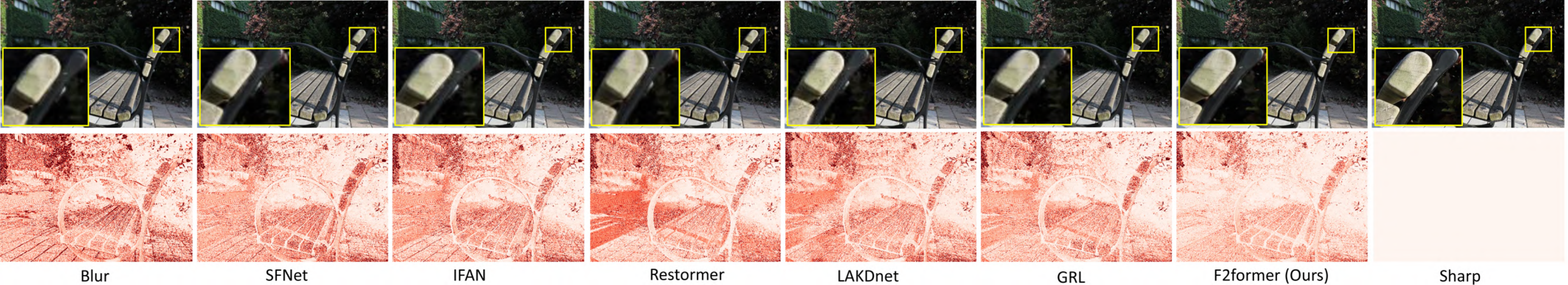}
    \caption{Visual illustration on the dual pixel DPDD test dataset. First row shows blurred images, predicted images of various methods, and GT sharp images. Second row shows the residual of the blurred image and predicted sharp images with GT. }
    \label{result:ddpd}
\end{figure*}

\begin{figure}
    \centering
    \includegraphics[width=\linewidth]{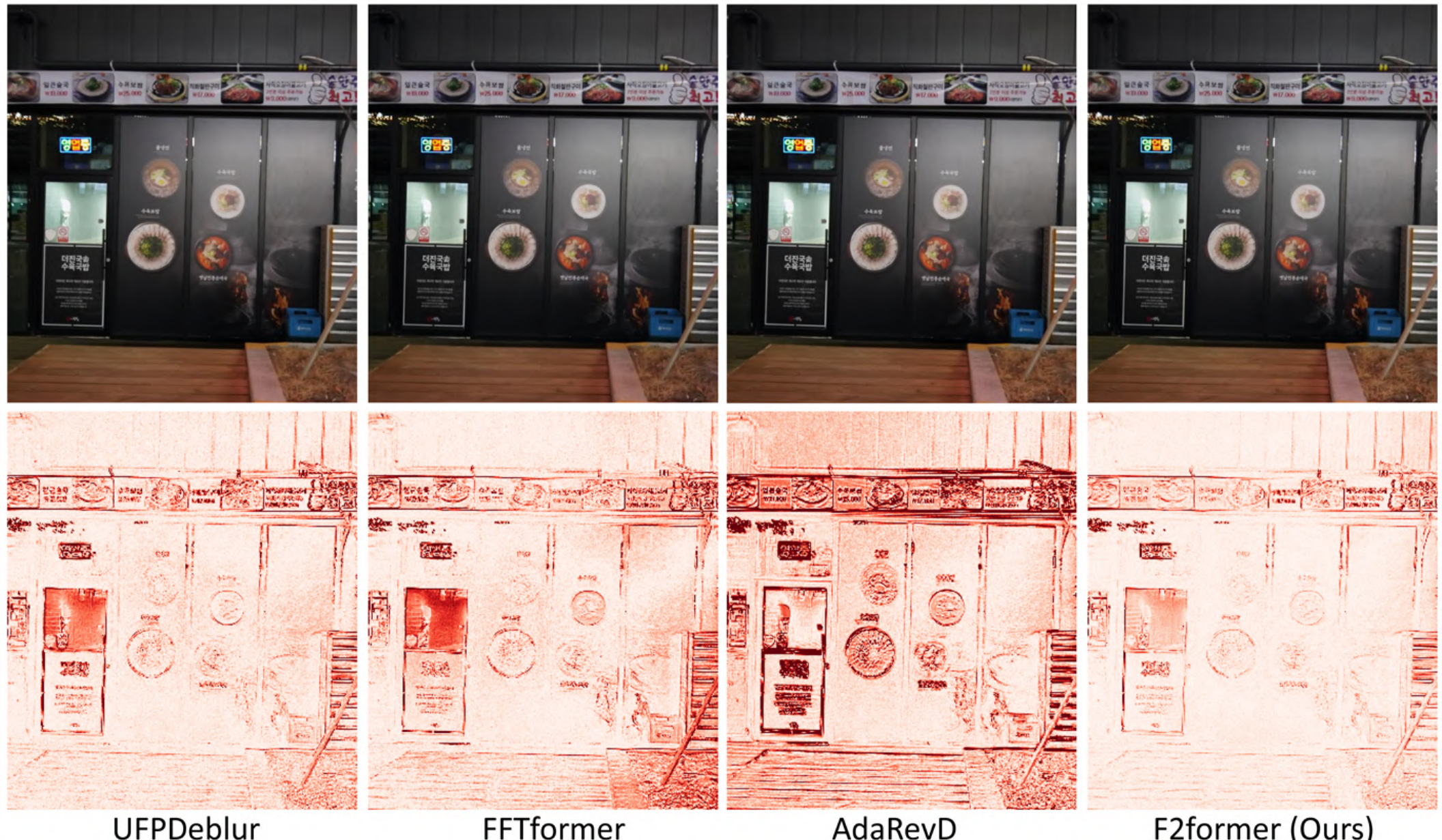}
    \caption{Visual illustration on RealBlur-J test dataset}
    \label{result:realblur}
\end{figure}

\subsection{Evaluation of trained model on GoPro}

Table \ref{tab:go-pro} shows the quantitative performance of F2former trained on the GoPro dataset and compares it with other SOTA methods on the test set of GoPro, HIDE, RealBlur-J and RealBlur-R datasets. We define this set-up as setting $\mathcal{A}$. As shown, F2former outperforms other methods in terms of both PSNR and SSIM for all four datasets. Compared to the second best, our base model achieves 0.53 dB gain in PSNR for the GoPro test dataset, while for the HIDE and RealBlur-J datasets, the gain is more than 1 dB. This can be validated from the visual comparison as shown in Figure \ref{result:go_pro}. The superior performance is attributed to the improved feature domain deblurring of F2WD and the representation power of FRFT-based transformer block. 

Table \ref{tab:go-pro} shows the parameter dependency and execution time of F2former. Since FRFT involves chirp multiplication and does not have an optimized implementation like FFT, the average runtime of F2former is slightly higher but competitive against other SOTA methods. In terms of parameters, the small model has 50\% less parameters with a notable performance drop, while the large model has marginal improvement with 42\% more parameters. Hence, the rest of the experiments are carried out with the base model.

\subsection{Evaluation of trained model on DPDD}
Table \ref{tab:ddpd} shows the performance comparison for different scenarios of defocus blur on the DPDD test dataset. We define this set-up as setting $\mathcal{B}$. In this case, the F2former is trained separately for both single and dual-pixel defocus deblurring tasks. Clearly, F2former has significant performance gains for all scene categories due to its ability to handle spatially varying blur. Specifically, for both single and dual pixel defocus deblurring cases, F2former achieves more than 2 dB performance gain over GRL \cite{grl}. This also can be validated from Figure \ref{result:ddpd}, where, the F2former removes the non-uniform blur most effectively.
\subsection{Evaluation of trained model on RealBlur}
\begin{table}
    \centering
    \caption{Comparison on RealBlur for setting $\mathcal{C}$.}
    \begin{adjustbox}{max width=\linewidth}
    \begin{tabular}{lcccccc}
        \toprule
        \multirow{2}{*}{Method} & \multicolumn{2}{c}{RealBlur-R} & \multicolumn{2}{c}{RealBlur-J} & \multicolumn{2}{c}{Average} \\
        \cmidrule(r){2-3} \cmidrule(r){4-5} \cmidrule(r){6-7}
        & PSNR $\uparrow$ & SSIM $\uparrow$ & PSNR $\uparrow$ & SSIM $\uparrow$ & PSNR $\uparrow$ & SSIM $\uparrow$ \\
        \midrule
        DeblurGAN-v2 & 36.44 & 0.935 & 29.69 & 0.870 & 33.07 & 0.903 \\
        SRN & 38.65 & 0.965 & 31.38 & 0.909 & 35.02 & 0.937 \\
        MPRNet & 39.31 & 0.972 & 31.76 & 0.922 & 35.54 & 0.947 \\
        MAXIM & 39.45 & 0.962 & 32.84 & 0.935 & 36.15 & 0.949 \\
        Stripformer & 39.84 & 0.974 & 32.48 & 0.929 & 36.16 & 0.952 \\
        DeepRFT+ & 40.01 & 0.973 & 32.63 & 0.933 & 36.32 & 0.953 \\
        FFTFormer & 40.11 & 0.975 & 32.62 & 0.933 & 36.37 & 0.954 \\
        UFPNet & 40.61 & 0.974 & 33.35 & 0.934 & 36.98 & 0.954 \\
        MRLPFNet & 40.92 & 0.975 & 33.19 & 0.936 & 37.06 & 0.956 \\
        AdaRevD & \underline{41.09} & \underline{0.978} & \underline{33.84} & \underline{0.943} & \underline{37.47} & \underline{0.961} \\
        F2former & \textbf{41.62} & \textbf{0.982} & \textbf{34.53} & \textbf{0.955} & \textbf{38.08} & \textbf{0.969} \\
        \bottomrule
    \end{tabular}
    \end{adjustbox}
    \label{tab:realblur}
\end{table}

We train our model on RealBlur-J and RealBlur-R separately. We define this set-up as setting $\mathcal{C}$ for which table \ref{tab:realblur} shows the performance comparison on respective test datasets. F2former has the best performance with more than 0.5-0.6 dB gain in PSNR compared to the second best. This also can be visualized from Figure \ref{result:realblur}, where F2former predicts sharper images compared to other SOTA methods. More visual results are provided in the supplementary.
\begin{figure}
    \centering
    \includegraphics[width=\linewidth]{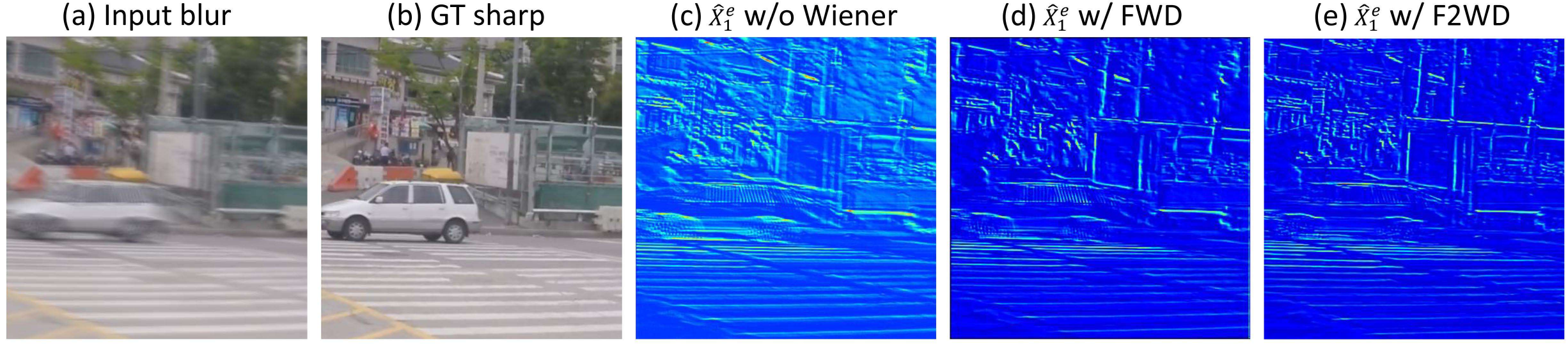}
    \caption{Visualization of shallow features to show the effect of F2WD (better visualized at 200\%)}
    \label{fig:ablation}
\end{figure}
\section{Analysis and Discussion}
We discuss the effect of each of the proposed module. For the related ablation studies, all experiments are performed with the base model trained with the GoPro dataset.\par
\textbf{Effect of F2WD.} The major contribution of F2WD is to refine shallow-level features by performing fractional Wiener deconvolution in feature space to produce sharper features which make the image reconstruction task efficient. Figure \ref{fig:ablation} supports the observation that applying Wiener deconvolution results in sharper features during training of F2former. Compared to feature-based Wiener deconvolution (FWD), the feature enhancement in the case of F2WD is sharper which also justifies our results in Table \ref{tab:my-table}. Introducing F2WD, there is 0.4 dB and 0.14 dB improvement in PSNR compared to baseline (no Wiener deconvolution) and the FWD case. to further verify, we retrain the Restormer with FWD and F2WD applied to shallow features before the transformer blocks. As shown in Table \ref{tab:my-table}, we see similar improvement compared to baseline Restormer performance.\par
\begin{figure}
    \centering
    \includegraphics[width=\linewidth]{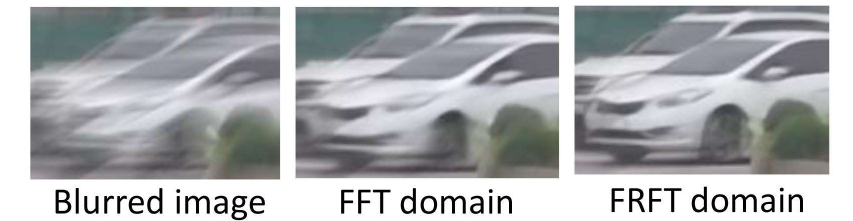}
    \caption{Effectiveness of FRFT-based self-attention on predicted image, compared to FFT based self-attention}
    \label{fig:frft_abl}
    \vspace{-0.3cm}
\end{figure}
\textbf{Effect of F2SA.} We implement FRFT from \cite{trainableFRFT} which has a complexity of $\mathcal{O}(N\log N)$, implemented based on \cite{frft-digital}. Hence, similar to FFTformer, F2SA has a space and time complexity of $\mathcal{O}(N)$ and $\mathcal{O}(NC\log N)$, which is lower compared to $\mathcal{O}(N^2)$ and $\mathcal{O}(N^2C)$ in conventional SA. Besides computational efficiency, as F2SA is based on FRFT, the attention computation occurs in the spatial-frequency domain resulting in capturing relevant information in both spatial and frequency domain. This helps in more effective deblurring as shown in Figure \ref{fig:frft_abl} where we replace FRFT with FFT operation during self-attention (SA) estimation. We also validate the effectiveness of phase modulation (PM) during SA computation by retraining FFTformer with this module. As shown in Table \ref{tab:my-table}, it improves its baseline performance by 0.17 dB in PSNR.\par
\begin{table}
\centering
\caption{Quantitative evaluation of each proposed component for different models}
\label{tab:my-table}
\begin{adjustbox}{max width=\linewidth}
\begin{tabular}{c|ccccc|c}
\hline
Models                     & w/ FWD & w/ F2WD & w/ PM in SA & w/ FM-FFN & w/ DFFN & PSNR/SSIM   \\ \hline
\multirow{2}{*}{Restormer} &  \cmark      &  \xmark       &   \xmark          &   \xmark        &  \xmark       & 33.48/0.964 \\
                           &   \xmark     &   \cmark       &   \xmark          &   \xmark        &  \xmark       & 33.63/0.969 \\ \hline
\multirow{2}{*}{FFTformer} &  \xmark      &   \xmark      &      \cmark         &    \xmark       &    \cmark       & 34.38/0.968 \\
                           &   \xmark     &    \xmark     &     \cmark         &    \cmark        &   \xmark      & 34.52/0.969 \\ \hline
\multirow{4}{*}{F2former}  &   \xmark     &    \xmark     &     \cmark        &     \xmark      &     \cmark    & 34.53/0.961 \\
                           &  \xmark       &     \xmark    &    \cmark          &     \cmark       &    \xmark     & 34.76/0.963 \\
                           &    \cmark    &  \xmark       &    \cmark          &   \cmark         &    \xmark     & 35.03/0.968 \\
                           &   \xmark      &   \cmark      &    \cmark          &   \cmark         &    \xmark     & 35.17/0.970 \\ \hline
\end{tabular}
\end{adjustbox}
\vspace{-0.4cm}
\end{table}

\textbf{Effect of FM-FFN.} We introduce frequency division multiplexing (FM) in the FFN block to emphasize significant frequency components. For baseline comparison, we replace our FM-FFN with discriminative frequency domain based FFN (DFFN) from \cite{fftformer} in our framework. As shown in Table \ref{tab:my-table}, there is a 0.22 dB PSNR reduction compared to using FM-FFN. For further validation, we include FM-FFN in FFTformer to retrain the model in the presence of PM in SA and notice an increase in performance by 0.14 dB PSNR. Further ablation studies related to choosing a cosine bell as a filter as well as visualization of its learnable threshold, the low and high pass response and also the $\alpha$ parameter for FRFT are carried out in Appendix C of supp.
\section{Conclusion}
In this work, we propose F2former, an effective deblurring approach based on the Fractional Fourier Transform (FRFT). As evident from the experiments, our Fractional-Fourier Transform (FRFT) based method achieves impressive deblurring even with spatially varying blur which is usually not handled well by spatial-only or Fourier-only methods. The main performance of our method comes from specially designed Fractional Feature based Wiener Deconvolution block and the Fractional Frequency aware Transformer block, which work with relevant spatio-spectral features for effective deblurring. The transformer layer extracts and utilizes relevant frequency components via a phase modulated SA and a frequency division multiplexing based feed-forward network. Experiments on different benchmark datasets for both motion and defocus blur show that our method performs favourably against other SOTA methods.\\
\textbf{Limitation:} F2former relies on the accuracy of estimated blur kernel by KEB, a separate trained model. Here, we also assume the CNN operations to be locally linear to apply F2WD which may result in erroneous feature maps occasionally. In future, we will investigate the blur kernel estimation in feature space only and incorporate it with the main deblurring network to train in an end-to-end manner for efficient image deblurring. We will also validate the effectiveness of FRFT for other image restoration tasks like super-resolution, denosining and dehazing.

{\small
\bibliographystyle{ieee_fullname}
\bibliography{egbib}
}
\section{Supplementary}
\setcounter{section}{0}
\renewcommand\thesection{\Alph{section}}

\section{Implementation details}
\subsection{Loss function}
For training F2former, we adopt L1 loss in both spatial and FFT domain similar to \cite{sfnet,xintm2024AdaRevD}. We also include L1 loss in FRFT domain to ensure intactness of the spatial-frequency information. Hence, the overall objective function to optimize our model is defined as,
\begin{equation}
    \mathcal{L}_{total} = \mathcal{L}_{s} + \lambda_{t_1}\mathcal{L}_{t_1} + \lambda_{t_{\alpha}}\mathcal{L}_{t_{\alpha}},
\end{equation}
where, $\mathcal{L}_{s}$, $\mathcal{L}_{t_1}$, and $\mathcal{L}_{t_{\alpha}}$ correspond to loss in spatial domain, FFT domain and FRFT domain with parameter $\alpha$, respectively. From Figure 3 (a) of main paper, they can be defined as,
\begin{equation}
    \begin{aligned}
        &\mathcal{L}_{s} = \sum_{p=1}^3 \frac{1}{N_p} ||\hat{\mathbf{Y}}_p - \mathbf{Y}_p||_1,\\
        &\mathcal{L}_{t_1} = \sum_{p=1}^3 \frac{1}{N_p} ||\mathcal{F}(\hat{\mathbf{Y}}_p) - \mathcal{F}(\mathbf{Y}_p)||_1,\\
        &\mathcal{L}_{t_{\alpha}} = \sum_{p=1}^3 \frac{1}{N_p} ||\mathcal{F}_{\alpha}(\hat{\mathbf{Y}}_p) - \mathcal{F}_{\alpha}(\mathbf{Y}_p)||_1,
    \end{aligned}
\end{equation}
where, $\mathbf{Y}_p$ is the scaled GT at $p$-th scale, $N_p$ is number of elements for normalization at $p$-th scale, $\mathcal{F}(\cdot)$ is FFT operation and $\mathcal{F}_{\alpha}(\cdot)$ is FRFT operation with parameter $\alpha$. $\lambda_{t_1}$ and $\lambda_{t_{\alpha}}$ are weights corresponding to $\mathcal{L}_{t_1}$ and $\mathcal{L}_{t_{\alpha}}$, respectively. We empirically decide $\mathcal{L}_{t_1} = \mathcal{L}_{t_{\alpha}} = 0.1$ and $\alpha = 0.5$.
\subsection{Parameter settings for training F2former}
We adopt the training strategy of NAFNet \cite{nafnet} and FFTformer \cite{fftformer}. Specifically, adopting their data augmentation strategy and Adam optimizer with default values. Initial learning rate is set at $1\times 10^{-3}$ and is updated with cosine annealing upto 600,000 iterations. The minimum learning rate is set at $1\times 10^{-7}$. F2former is trained with input images of shape $256\times 256 \times 3$ with a batch size of 8. For computing self-attention we set the patch size as $8\times 8$. Other parameters like $\alpha$ for FRFT computation in FHTB block and $\lambda$ as the cut-off frequency for cosine-bell function in FM-FFN are set as learnable parameter and are updated during training based on minimizing the loss function. The corresponding details are discussed later in ablation section. All the experiments are carried out in a single 80GB Nvidia A100 GPU.

\subsection{Kernel estimation block (KEB)}
\begin{figure}
    \centering
    \includegraphics[width=\linewidth]{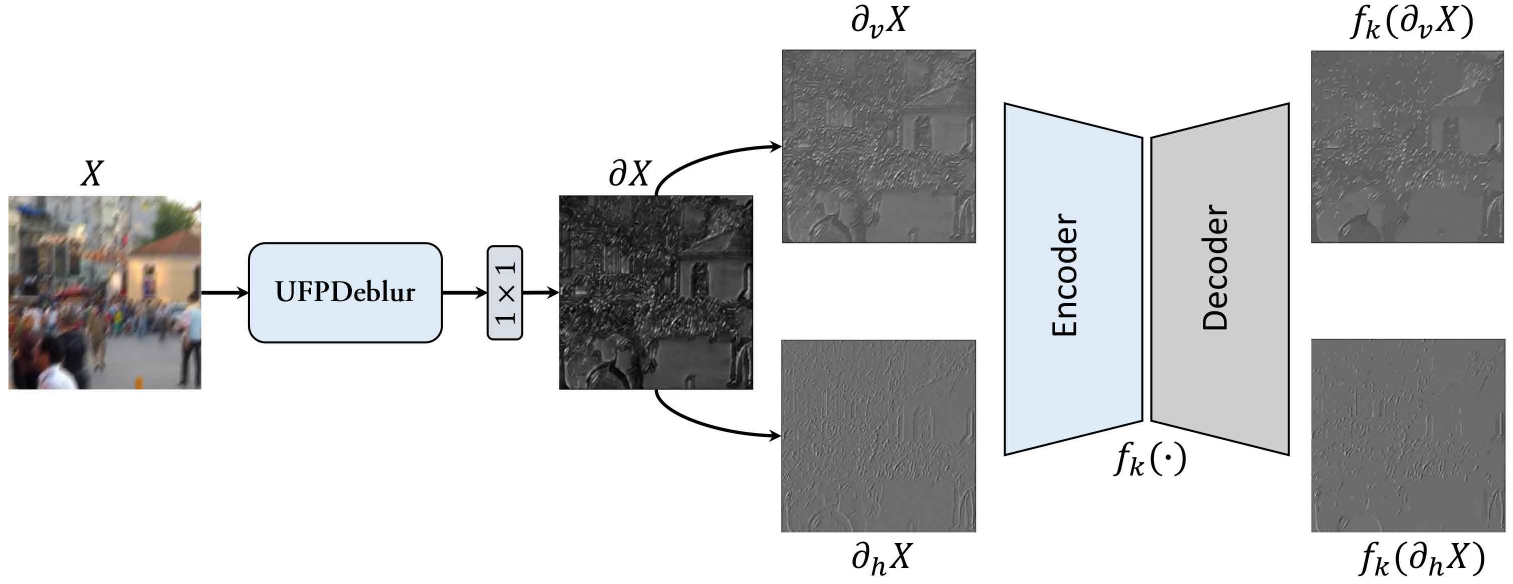}
    \caption{Workflow of our KEB module.}
    \label{fig:keb}
\end{figure}

For a given blurry input $\mathbf{X}$, KEB estimates its corresponding blurry kernel $\mathbf{K}$. Its detail overflow is shown in Figure \ref{fig:keb}. Primarily, we use kernel estimation network from \cite{ufpdeblur} where for each pixel the blur kernel is estimated of shape $19\times 19$. However, this kernel can not be used directly for Wiener deconvolution as it will increase the overall computation cost. Therefore, upon reshaping this estimated kernel, it turns out the gradient map $(\partial \mathbf{X})$ while also highlighting the blur pattern. To estimate an accurate blur kernel from $\partial \mathbf{X}$, we leverage the blur kernel estimation method in \cite{keb}. Basically, we estimate gradient at the vertical $(\partial_v \mathbf{X})$ and horizontal $(\partial_h \mathbf{X})$ direction from $\partial \mathbf{X}$ which is later passed to a 6-layer auto-encoder as shown in Figure \ref{fig:keb}. The encoder part removes unnecessary details of image gradients while the decoder extracts the major structural details with enhanced edges formulated as $\partial_v \mathbf{X}_e = f_K(\partial_v \mathbf{X})$ and $\partial_h \mathbf{X}_e = f_K(\partial_h \mathbf{X})$, where $f_K(\cdot)$ is auto-encoder operation as shown in above Figure. The design of auto-encoder as well as its training is carried out according to the respective details in \cite{keb}.\par
After obtaining the salient edges as $\partial \mathbf{X}_e = f_K(\partial \mathbf{X})$ for blurry input $\mathbf{X}$, the blur kernel $(\mathbf{K})$ can be estimated from the following optimization problems,
\begin{equation}\label{eq17}
\begin{aligned}
    &\mathbf{K} = \arg \min_\mathbf{K} ||\mathbf{K}\cdot\partial \mathbf{X}_e - \partial \mathbf{X}||_2^2 + \eta ||\mathbf{K}||_2^2,\\
    &\mathbf{X}_l = \arg \min_{\mathbf{X}_l} ||\mathbf{K}\cdot \mathbf{X}_l - \mathbf{X}||_2^2 + \gamma ||\mathbf{X}_l||_2^2,
\end{aligned}
\end{equation}
where, $\eta$ and $\gamma$ are parameters for regularization terms, and $\mathbf{X}_l$ is intermediate latent image. Similar to \cite{keb}, we use $\mathbf{X}_l$ to update the edge information in $\partial \mathbf{X}_e$. From \cite{keb}, the closed form solution of the first optimization problem in equation \ref{eq17} can be written in continuous domain as,
\begin{equation}\label{eq18}
    K = \mathcal{F}^{-1}\left( \frac{\overline{\mathcal{F}(\partial_h {X}_e)}\mathcal{F}(\partial_h {X}+\overline{\mathcal{F}(\partial_v {X}_e)}\mathcal{F}(\partial_v {X})}{\mathcal{F}(\partial_h {X}_e)^2 + \mathcal{F}(\partial_h {X}_e)^2 + \eta}\right),
\end{equation}
where $\mathcal{F}^{-1}(\cdot)$ is inverse FFT and $\overline{\mathcal{F}(\cdot)}$ is complex conjugate operation FFT domain. Then the kernel estimation can be formulated as per following algorithm
\begin{figure}
    \centering
    \includegraphics[width=\linewidth]{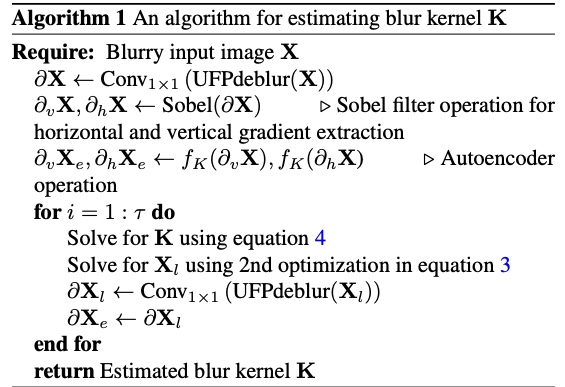}
    \label{fig:enter-label}
\end{figure}

\begin{figure}
    \centering
    \includegraphics[width=\linewidth]{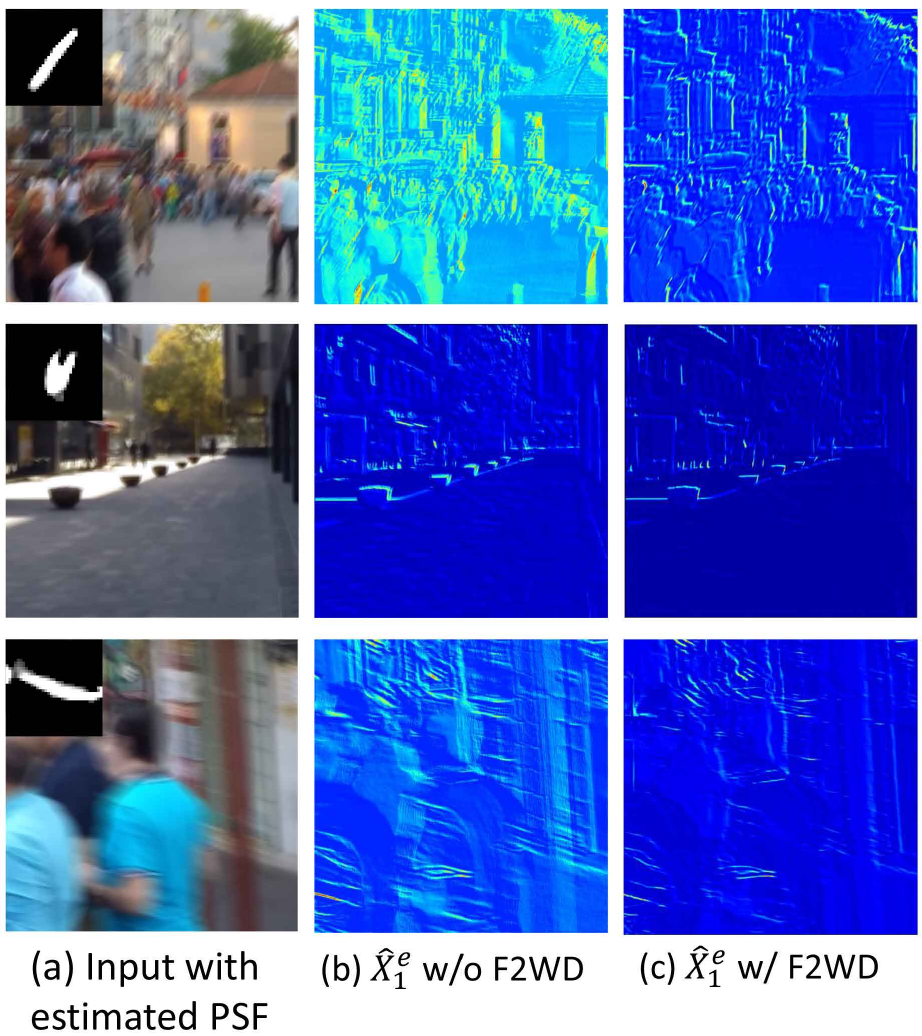}
    \caption{Estimated Kernel for a given blurry image and visualization of F2WD operation using it. It is also compared with features without any Wiener deconvolution operation for proper understanding.}
    \label{fig:keb-example}
\end{figure}

Here, the values of parameters $\eta$, $\gamma$, and $\tau$ are selected as 1, 0.002, and 15, respectively, according to \cite{keb}. Figure \ref{fig:keb-example} shows the estimated kernel using the above algorithm as well as F2WD operation using this. Clearly, using the estimated kernel, F2WD is able to perform deblurring in latent space accurately. The accuracy of F2WD can be further improved if more accurate kernel estimation algorithm can be developed in feature space itself. We will investigate this further in future studies.

\section{Proof of Proposition 1}
The signal observation model in image domain can be written as,
\begin{equation}\label{eq21}
    \mathbf{Y} = \mathbf{K}\mathbf{X} + \mathbf{n},
\end{equation}
where, $\mathbf{Y}$ is observed data, $\mathbf{X}$ is true signal, $\mathbf{n}$ is associated noise elements and $\mathbf{K}$ is the degradation process. Here, we consider the noise $\mathbf{n}$ to be independent of input processes and is zero mean.

We assume that the Wiener deconvolution operator is $\mathbf{G}$ in FRFT domain. Then, we can define the estimate of the true signal as,
\begin{equation}\label{eq20}
    \mathbf{\hat{X}} = \mathbf{F}^{-\alpha}\mathbf{C}\mathbf{G}\mathbf{F}^{\alpha}\mathbf{Y},
\end{equation}
where, $\mathbf{C}$ corresponds to chirp multiplication with with diagonal elements $\mathbf{C}_{kk} = e^{-ik^2\cot \theta/2}$. This is included due to the convolution theorem in FRFT domain as shown in equation 8 (main paper). The design of filter or Wiener deconvolution operator is based on minimizing the mean square error (MSE), which can be defined as following,
\begin{equation}
    J_\alpha = \frac{1}{N} \mathbb{E}\{(\mathbf{X}-\mathbf{\hat{X}})^H (\mathbf{X}-\mathbf{\hat{X}}) \}
\end{equation}
Now, substituting equation \ref{eq20} in above equation will result in 
\begin{equation}\label{eq22}
    \begin{aligned}
        J_\alpha &= \frac{1}{N} \mathbb{E}\{(\mathbf{X}-\mathbf{F}^{-\alpha}\mathbf{C}\mathbf{G}\mathbf{F}^{\alpha}\mathbf{Y})^H (\mathbf{X}-\mathbf{F}^{-\alpha}\mathbf{C}\mathbf{G}\mathbf{F}^{\alpha}\mathbf{Y}) \}\\
        & = \frac{1}{N} \left[\mathbb{E}\{\mathbf{X}^H\mathbf{X}\} - \mathbb{E}\{\mathbf{X}^H\mathbf{F}^{-\alpha}\mathbf{C}\mathbf{G}\mathbf{F}^{\alpha}\mathbf{Y}\}  \right.\\
        &\left. - \mathbb{E}\{(\mathbf{F}^{-\alpha}\mathbf{C}\mathbf{G}\mathbf{F}^{\alpha}\mathbf{Y})^H\mathbf{X}\} \right.\\
        &\left.+ \mathbb{E}\{(\mathbf{F}^{-\alpha}\mathbf{C}\mathbf{G}\mathbf{F}^{\alpha}\mathbf{Y})^H \mathbf{F}^{-\alpha}\mathbf{C}\mathbf{G}\mathbf{F}^{\alpha}\mathbf{Y} \}\right]
    \end{aligned}
\end{equation}
Now, we analyse each term in equation \ref{eq22}. For the first term $\mathbb{E}\{\mathbf{X}^H\mathbf{X}\}$, it suggest the autocorrelation matrix of true signal denoted as $\mathbf{S}^{xx} = \mathbb{E}\{\mathbf{X}^H\mathbf{X}\}$.

The second term can be re-written in terms of cross-correlation matrix between $\mathbf{X}$ and $\mathbf{Y}$, which can be defined as, $\mathbf{S}^{xy} = \mathbb{E}\{\mathbf{X}\mathbf{Y}^H\}$. Hence, $\mathbb{E}\{\mathbf{X}^H\mathbf{F}^{-\alpha}\mathbf{C}\mathbf{G}\mathbf{F}^{\alpha}\mathbf{Y}\} = \mathbb{E}\{\mathbf{S}^{xy} \mathbf{F}^{-\alpha}\mathbf{C}\mathbf{G}\mathbf{F}^{\alpha}\}$.

Similarly the third term can be written as, 
\begin{equation}
    \mathbb{E}\{(\mathbf{F}^{-\alpha}\mathbf{C}\mathbf{G}\mathbf{F}^{\alpha}\mathbf{Y})^H\mathbf{X}\} = \mathbb{E}\{ (\mathbf{F}^{-\alpha}\mathbf{C}\mathbf{G}\mathbf{F}^{\alpha})^H\mathbf{S}^{yx}\}.
\end{equation}

For the last term, it can be simplified as following,
\begin{equation}\label{eq23}
\begin{aligned}
    &\mathbb{E}\{(\mathbf{F}^{-\alpha}\mathbf{C}\mathbf{G}\mathbf{F}^{\alpha}\mathbf{Y})^H \mathbf{F}^{-\alpha}\mathbf{C}\mathbf{G}\mathbf{F}^{\alpha}\mathbf{Y} \} \\
    &=\mathbb{E}\{\mathbf{Y}^H \mathbf{F}^{-\alpha} \mathbf{G}^H \mathbf{C}^H \mathbf{F}^{\alpha} \mathbf{F}^{-\alpha}\mathbf{C}\mathbf{G}\mathbf{F}^{\alpha}\mathbf{Y}\}\\
    &=\mathbb{E}\{\mathbf{Y}^H \mathbf{F}^{-\alpha} \mathbf{G}^H \mathbf{G}\mathbf{F}^{\alpha}\mathbf{Y}\}
\end{aligned}
\end{equation}
In equation \ref{eq23}, the last line is from the properties of FRFT that $\mathbf{F}^{\alpha}\mathbf{F}^{-\alpha} = \mathbf{F}^{-\alpha+\alpha} = \mathbf{F}^0$, which is a unitary matrix. Similarly, $\mathbf{C}^H\mathbf{C}=\mathbf{I}$, due to the orthogonal properties of chirp function.

\begin{figure*}
    \centering
    \includegraphics[width = \textwidth]{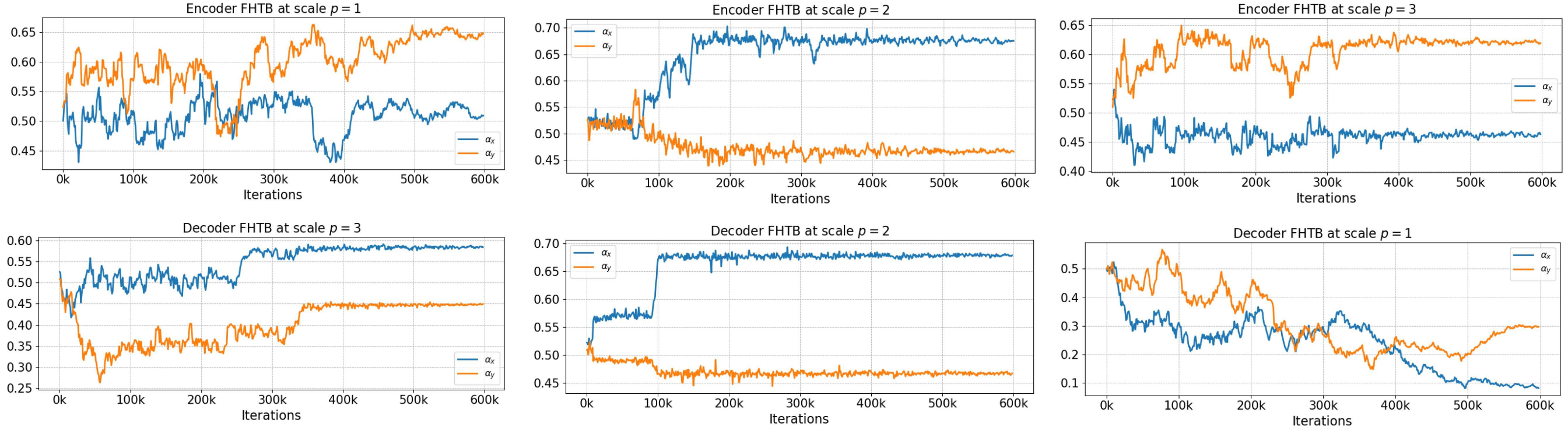}
    \caption{Visualization of variation of $\alpha$ for FRFT operations in FHTBs at different scales of F2former base model}
    \label{fig:alpha}
\end{figure*}

Now, as the $\mathbf{G}$ consists of only diagonal elements, the equation \ref{eq22} can be rewritten in terms of trace of the matrix. So, using the above expression, it can be reformulated as,
\begin{equation}
    \begin{aligned}
        J_\alpha &= \frac{1}{N} [\text{tr}(\mathbf{S}^{xx}) - \text{tr}(\mathbf{S}^{xy} \mathbf{F}^{-\alpha}\mathbf{C}\mathbf{G}\mathbf{F}^{\alpha}) \\
        & -\text{tr}((\mathbf{F}^{-\alpha}\mathbf{C}\mathbf{G}\mathbf{F}^{\alpha})^H\mathbf{S}^{yx}) + \text{tr}(\mathbf{Y}^H \mathbf{F}^{-\alpha} \mathbf{G}^H \mathbf{G}\mathbf{F}^{\alpha}\mathbf{Y})],
    \end{aligned}
\end{equation}
where, $\text{tr}(\cdot)$ is trace of matrix. Now, for each diagonal element of $\mathbf{G}_j$, we minimize $J_\alpha$ as following,
\begin{equation}
    \begin{aligned}
        \frac{\partial J_\alpha}{\partial \mathbf{G}_j} = - (\mathbf{C}_{jj}\mathbf{F}^{\alpha}_{jj}\mathbf{S}^{xy}_j\mathbf{F}^{-\alpha}_{jj})^H   +\mathbf{F}^{\alpha}_{jj} \mathbf{S}^{yy}_j \mathbf{F}^{-\alpha}_{jj} \mathbf{G}_j.
    \end{aligned}
\end{equation}
To minimize $J_\alpha$, we have to make above equation 0, and then solving for $\mathbf{G}_j$ we get, 
\begin{equation}\label{final}
    \mathbf{G}_j = \frac{(\mathbf{C}_{jj}\mathbf{F}^{\alpha}_{jj}\mathbf{S}^{xy}_j\mathbf{F}^{-\alpha}_{jj})^H}{\mathbf{F}^{\alpha}_{jj} \mathbf{S}^{yy}_j \mathbf{F}^{-\alpha}_{jj}},
\end{equation}
where, $\mathbf{S}^{yy}$ is auto-correlation of signal $\mathbf{Y}$, which can be defined as,
\begin{equation}\label{28}
\begin{aligned}
    \mathbf{S}^{yy} &= \mathbb{E}\{\mathbf{Y}\mathbf{Y}^H\} = \mathbb{E}\{(\mathbf{K}\mathbf{X}+\mathbf{n})(\mathbf{K}\mathbf{X}+\mathbf{n})^H\}\\
    & = \mathbb{E}\{\mathbf{K}\mathbf{X}\mathbf{X}^H\mathbf{K}^H\} + \mathbb{E}\{\mathbf{n} \mathbf{n}^H\} + \mathbb{E}\{\mathbf{n}\mathbf{X}^H\mathbf{K}^H\} \\
    &+  \mathbb{E}\{\mathbf{X}\mathbf{K}\mathbf{n}^H\}. 
\end{aligned}
\end{equation}
In above equation, $\mathbb{E}\{\mathbf{n}\mathbf{X}^H\mathbf{K}^H\} =\mathbb{E}\{\mathbf{X}\mathbf{K}\mathbf{n}^H\}=0$, as $\mathbf{X}$ and $\mathbf{n}$ are independent. Now, as $\mathbf{S}^{nn} = \mathbb{E}\{\mathbf{n} \mathbf{n}^H\}$ and $\mathbf{S}^{xx} = \mathbb{E}\{\mathbf{X} \mathbf{X}^H\}$, we replace them in above equation to get 
\begin{equation}\label{eq28}
\mathbf{S}^{yy} = \mathbf{K}\mathbf{S}^{xx}\mathbf{K}^H + \mathbf{S}^{nn}.
\end{equation}
Similarly, the cross correlation term $\mathbf{S}^{xy}$ can be written as,
\begin{equation}\label{29}
\begin{aligned}
    \mathbf{S}^{xy} &=  \mathbb{E}\{\mathbf{X}\mathbf{Y}^H\} = \mathbb{E}\{\mathbf{X}(\mathbf{KX} + \mathbf{n})^H\}\\
    & = \mathbb{E}\{\mathbf{X}\mathbf{X}^H\mathbf{K}^H\} + \mathbb{E}\{\mathbf{X}\mathbf{n}^H\} = \mathbf{S}^{xx}\mathbf{K}^{H}. 
\end{aligned}
\end{equation}
Now transferring equation \ref{28} and \ref{29} in equation \ref{final}, we will get the derived formulation in equation 7 (main paper). Hence, this completes the proof.

\section{Ablation study (continued)}
\subsection{Effect of F3RB} We train our base model without F3RB which was added to extract features in terms of different contextual information with respect to varied $\alpha$ parameter for FRFT operation. We observe a 0.13 dB drop in PSNR compared to our baseline performance.
\subsection{Visualization of learnable $\alpha$ for FRFT in F2WD and FHTB}
In all FRFT operations in F2former, we have taken the $\alpha$ parameter as learnable parameter, specifically, we set $\alpha = \{\alpha_x, \alpha_y\}$ to apply FRFT in horizontal and vertical directions of an image separately. The main reason behind this is that based on different blur content at different scales there would be separate requirement of $\alpha$ parameters to represent optimal spatial-frequency information. It also depends on specific task of the block like the role of FHTB blocks in encoder side is different compared to that of decoder side (encoder part minimizes the blur content at different scales, whereas the decoder part efficiently reconstructs the sharp image details). This can be visualized from Figure \ref{fig:alpha} where each plot represent the average $\alpha$ variation with respect to training iterations and corresponding to a particular FHTB block at particular scale. Usually, for encoder operations $\alpha$ varies between (0.4, 0.7), whereas for decoder operation it varies between (0.1,0.7). Our hypothesis is that as decoder (specially at last stage) mainly focuses on reconstructing the sharp image details, it majorly focuses on the spatial content of the image making it low $\alpha$ values as shown in last plot of Figure \ref{fig:alpha}. It is also noticeable from the above figure that $\alpha$ also varies across horizontal and vertical directions. This is mostly due to focus on different important gradient information at separate scales and separate levels (encoder or decoder).

\begin{figure}
    \centering
    \includegraphics[width = \linewidth]{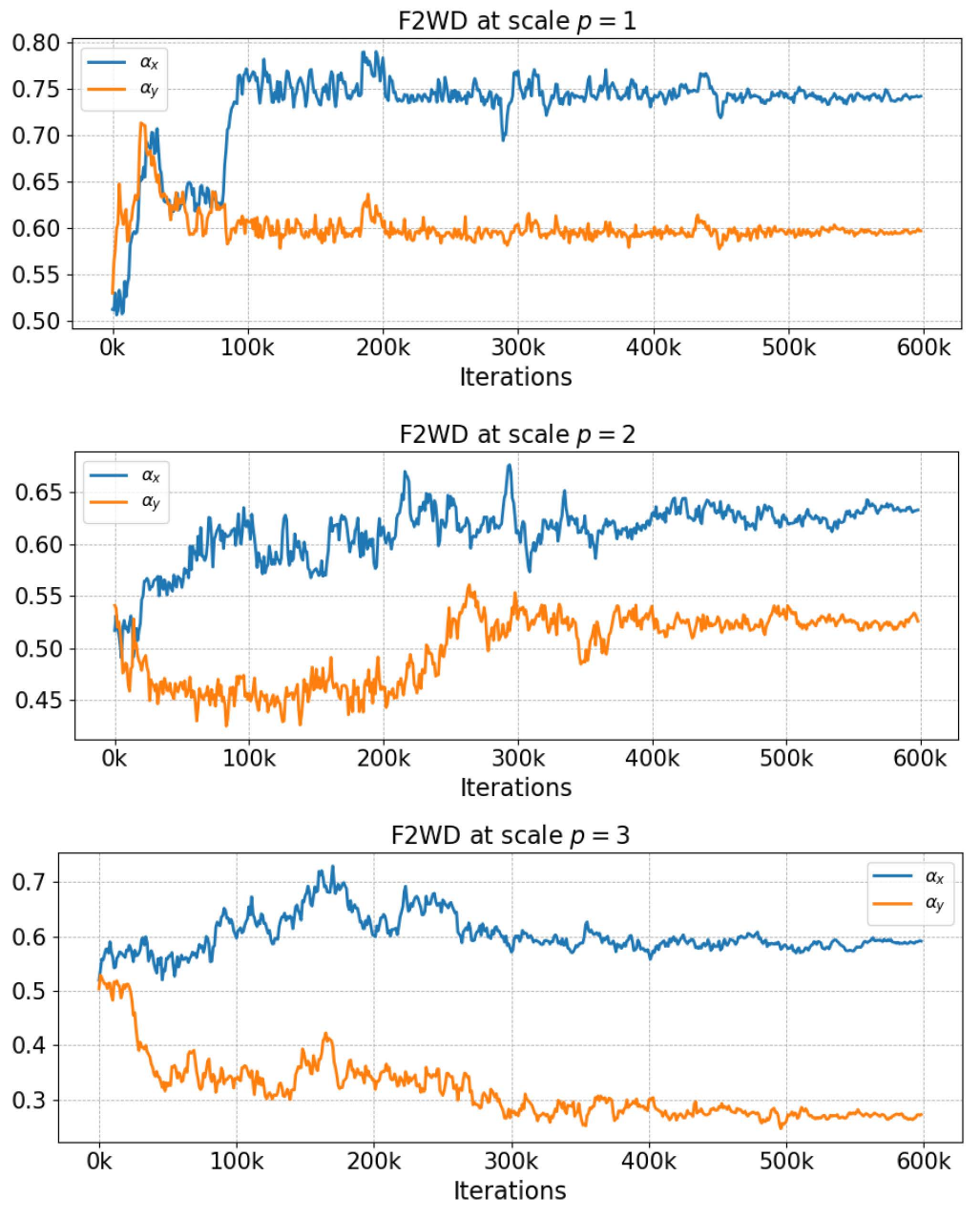}
    \caption{Visualization of variation of $\alpha$ for FRFT operations in F2WDs at different scales of F2former base model}
     \label{fig:alpha-vary2}
\end{figure}

Figure \ref{fig:alpha-vary2} shows the $\alpha$ variation to perform featured based fractional Wiener deconvolution at different scales. As at downsampled level, the spatially varying blur dominates more, we see more focus of F2WD to spatial content (low $\alpha$ values) as the model gets deeper.
Due to this optimal choice of $\alpha$, F2WD is able to perform feature based deblurring operation more effectively as shown in Figure \ref{fig:keb-example}. For further verification, we retrain our model with fixed $\alpha=0.5$ for both horizontal and vertical direction as suggested by \cite{deepFrac} for both F2WD and FSGT. We observe a 0.18 dB reduced PSNR compared to our baseline performance.
\subsection{Effect of Cosine-bell function in FM-FFN}
\begin{figure}
    \centering
    \includegraphics[width = \linewidth]{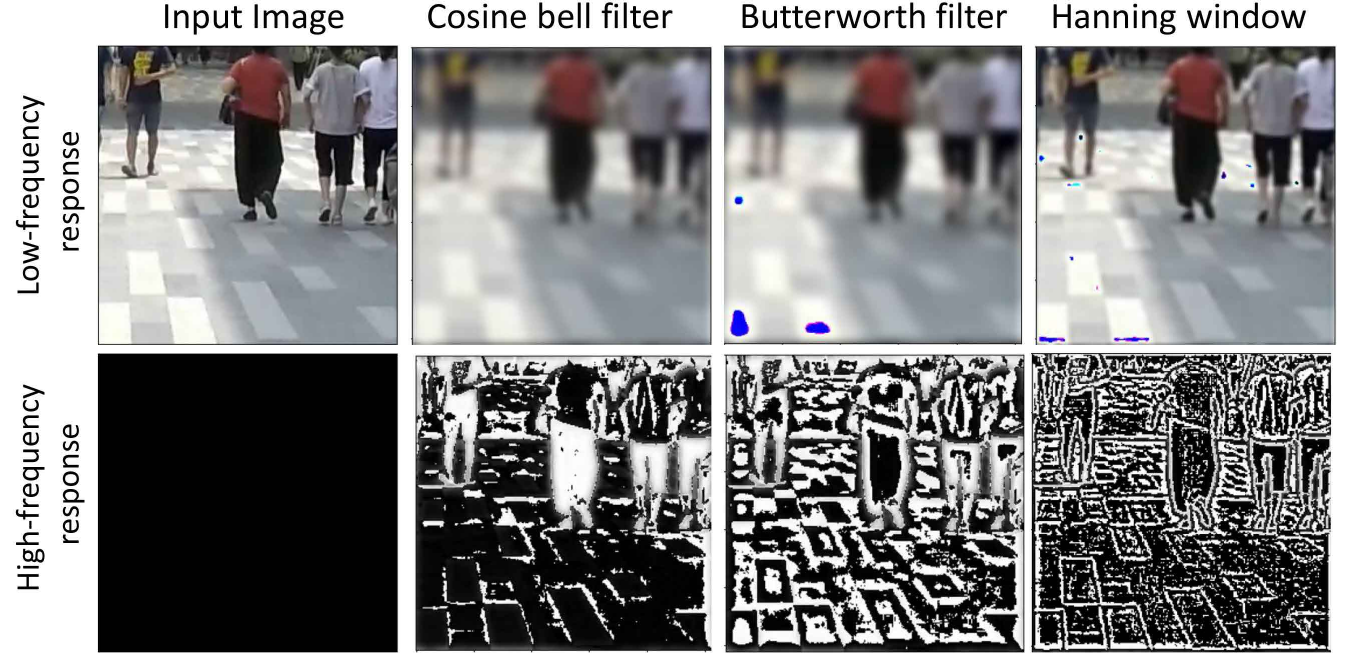}
    \caption{Visualization of extracted low-pass and high-pass image signal using different filters.}
    \vspace{0.25cm}
     \label{fig:filter_response}
\end{figure}

The cosine bell formulation in equation 13 (main paper) is generic. The exact formulation is defined as below,
\begin{equation}
    \mathbf{W}^l_{jk} = \begin{cases} 
  1 & \text{if } |u| \leq u_c - \frac{u_s}{2} \\ 
 0.5\left[1+ \cos \left(\frac{\pi(|u| - u_c + \frac{u_s}{2})}{u_s}\right) \right] & \text{if } \lambda_{ls}<|u| < \lambda_{rs}\\
 0 & \text{if } |u| \geq u_c + \frac{u_s}{2},
  \end{cases}
\end{equation}
where, $\lambda_{ls} = u_c - \frac{u_s}{2}$, $\lambda_{rs} = u_c + \frac{u_s}{2}$, and $u=\sqrt{j^2+k^2}$. So, threshold $\lambda$ in equation 13 is basically consists of two parameters, $u_c$ which is cut-off frequency and $u_s$ which is pass band frequency to reduce ringing effect in spatial domain. Figure \ref{fig:filter_response} shows the advantage of using cosine bell function compared to other formulations like butterworth and hanning window operation. Clearly cosine bell reduces the ringing artifacts and reconstructs high and low frequency images more effectively for same cut-off frequency.

\begin{figure}
    \centering
    \includegraphics[width = \linewidth]{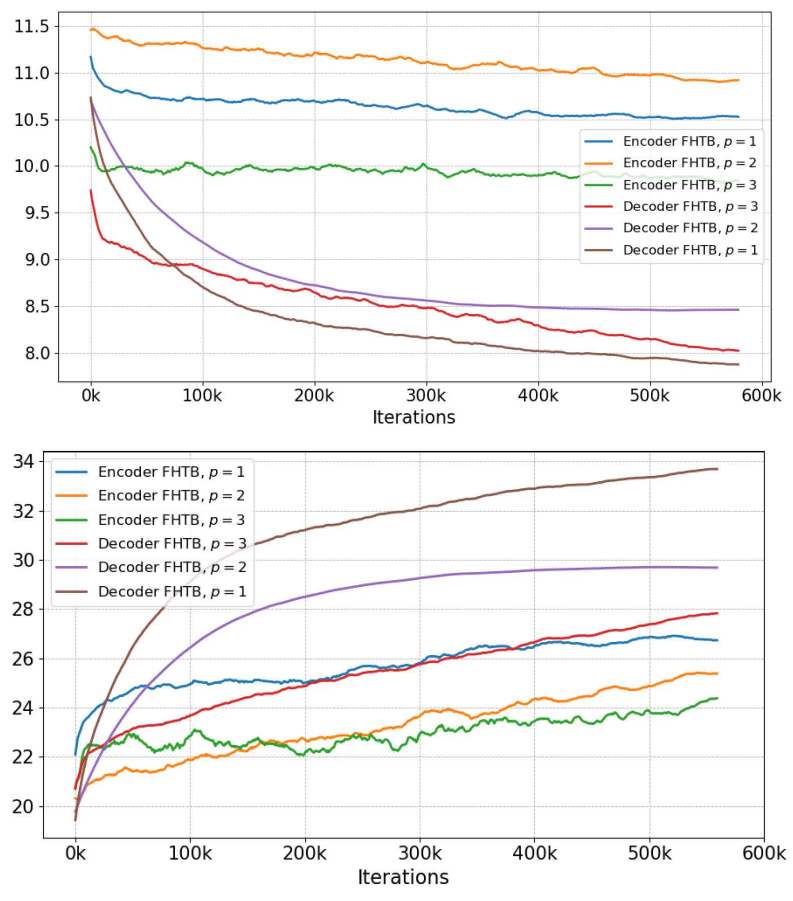}
    \caption{Variation of $u_c$ (\textbf{top}) and $u_s$(\textbf{bottom}) across the training iterations.}
     \label{fig:filter_graph}
\end{figure}

We keep both $u_c$ and $u_s$ as the learning parameter during training our network. Figure \ref{fig:filter_graph} shows the variation of $u_c$ and $u_s$ across the training iterations. For optimal high and low frequency extraction, the cut-off frequency for all FHTB blocks varies between (8,12) in saturation whereas to reduce the ringing effect the pass band frequency mainly varies in (24,34) across all the FHTBs. For further validation we use the three filters shown in Figure \ref{fig:filter_response} for training upto 200k iterations on GoPro dataset. The respective test results are shown in Table \ref{tab:my-table5}. Clearly, the utilization of cosine bell function as a filter benefits our proposed F2former compared to other filters.

\begin{table}[]
\centering
\caption{Effect of different filters on the performance of F2former}
\label{tab:my-table5}
\begin{tabular}{ccc}
\hline
Filters        & PSNR  & SSIM  \\ \hline
Cosine Bell    & 33.24 & 0.961 \\
Butterworth    & 33.02 & 0.953 \\
Hanning window & 33.11 & 0.957 \\ \hline
\end{tabular}
\end{table}

\subsection{Visualization of effect of FM-FFN}
\begin{figure}
    \centering
    \includegraphics[width = \linewidth]{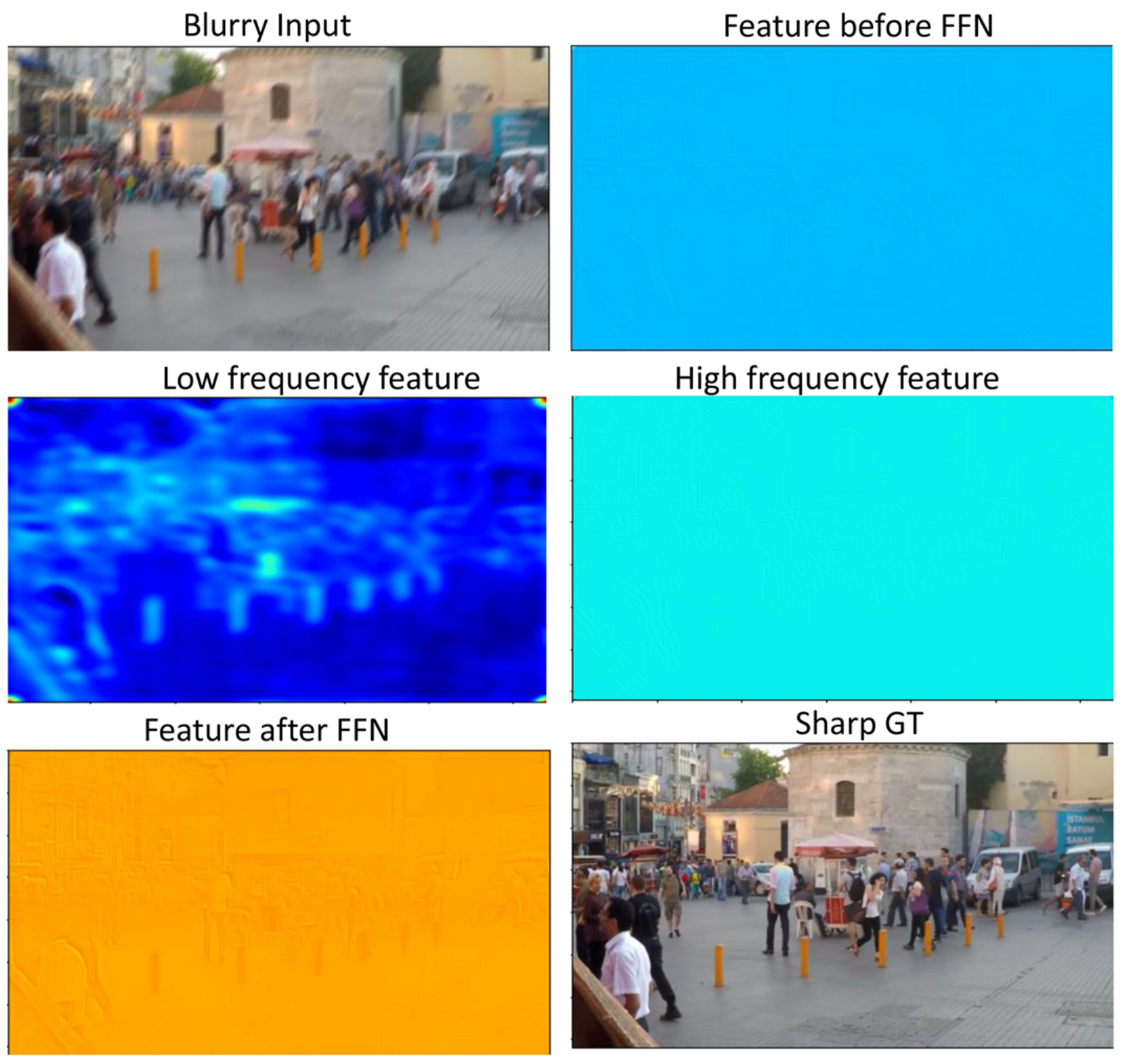}
    \caption{Visualization of extracted low and high frequency features in FM-FFN of last F2TB block (better visualized in 200-300\%)}
     \label{fig:ffn_response}
\end{figure}
Figure \ref{fig:ffn_response} shows how FM-FFN extracts high and low frequency features, and based on important frequency information it generates sharp  details as shown in the Figure. Clearly, compared to before FFN features, the features after FFN has more sharper edges and more structural details.

\begin{figure*}
    \centering
    \includegraphics[width=\textwidth]{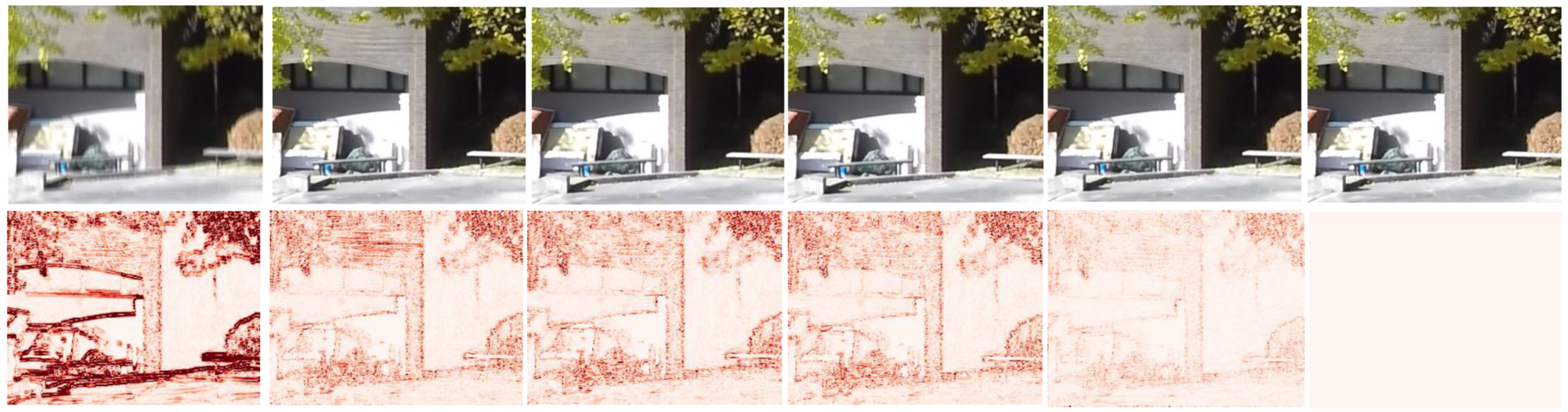}
    \includegraphics[width=\textwidth]{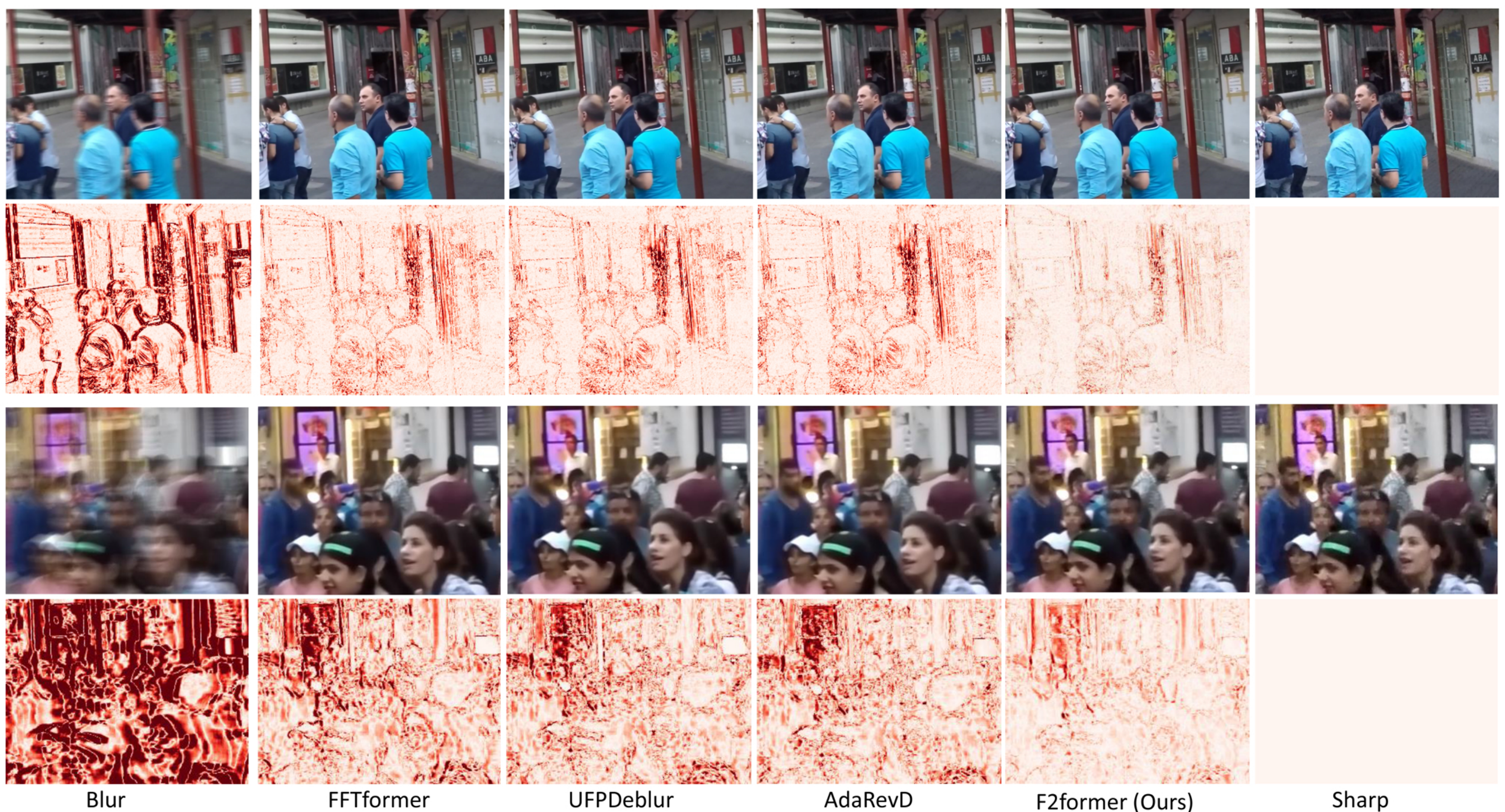}
    \caption{Examples on the GoPro test dataset. The odd rows show blur image, predicted images of different methods, and GT
sharp image. The even rows show the corresponding residual of the blur image / predicted sharp images and GT sharp image. Better visualized at 200\%.}
    \label{fig:gopro}
\end{figure*}

\begin{figure*}
    \centering
    \includegraphics[width=\textwidth]{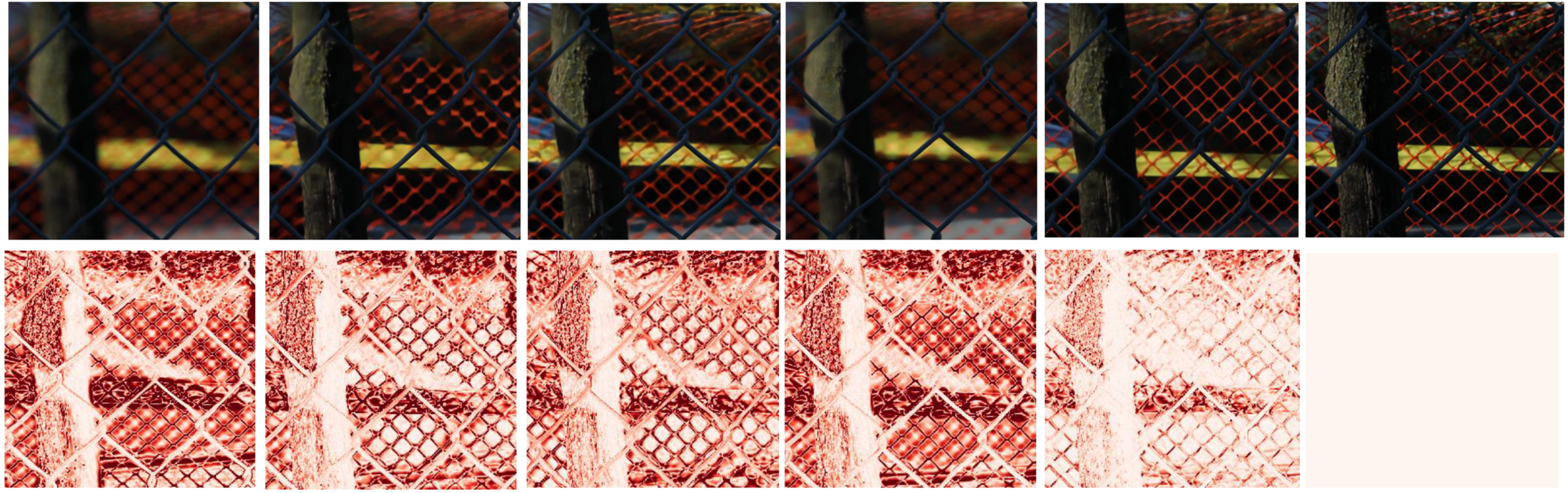}
    \includegraphics[width=\textwidth]{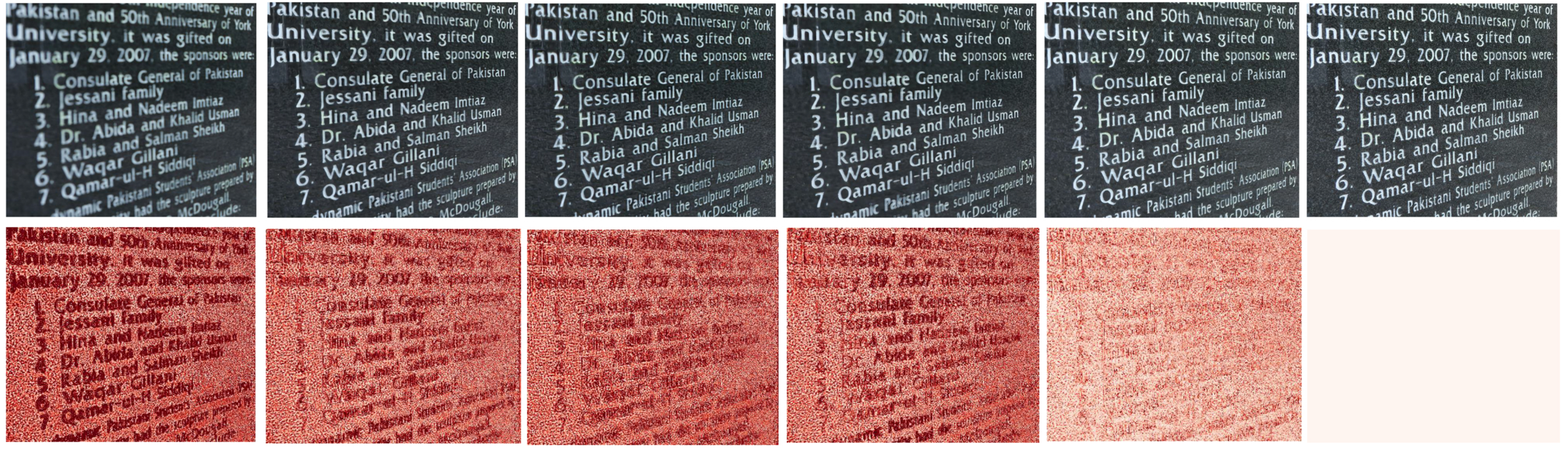}
    \includegraphics[width=\textwidth]{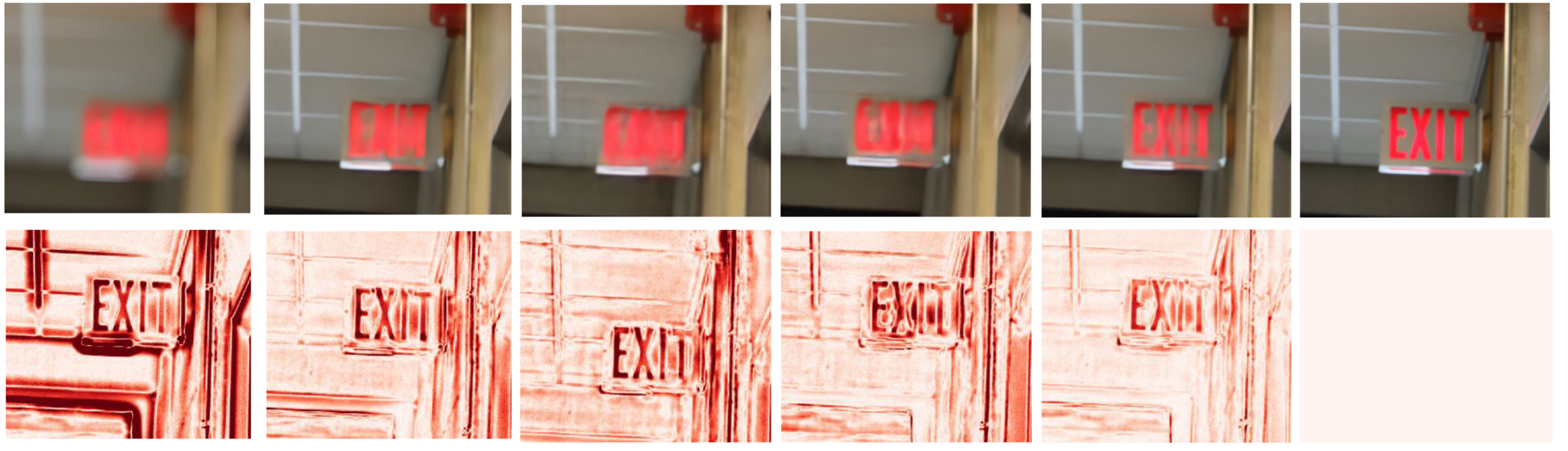}
    \includegraphics[width=\textwidth]{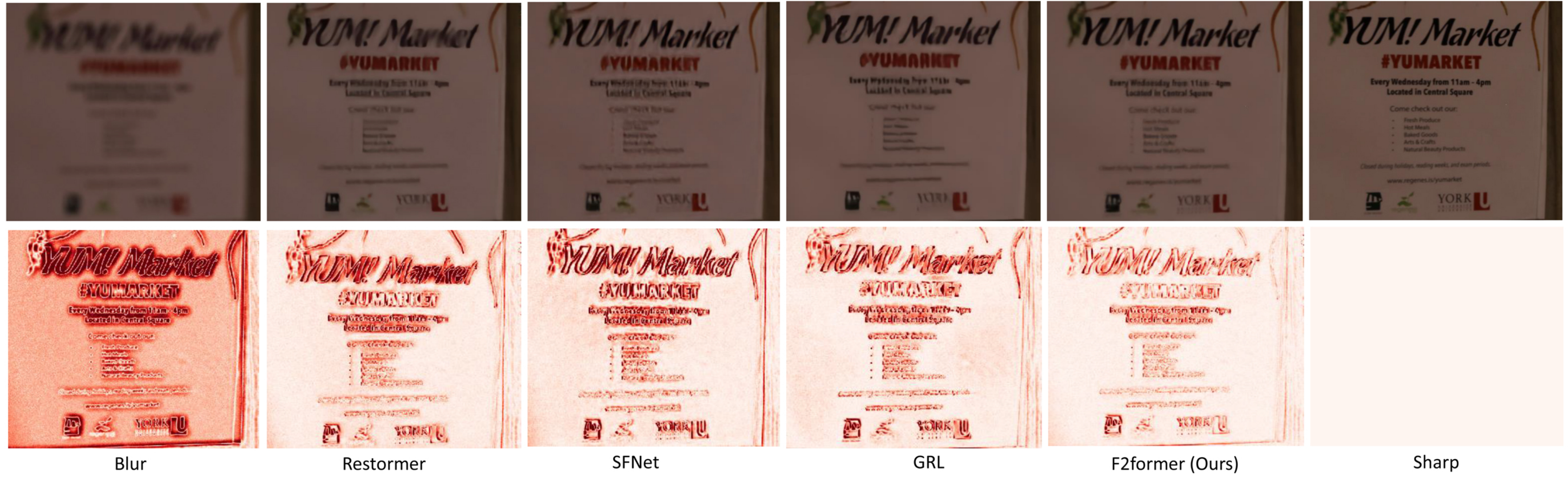}
    \caption{Examples on the DDPD-single pixel test dataset. The odd rows show blur image, predicted images of different methods, and GT
sharp image. The even rows show the corresponding residual of the blur image / predicted sharp images and GT sharp image. Better visualized at 200\%.}
    \label{fig:ddpd-s}
\end{figure*}

\begin{figure*}
    \centering
    \includegraphics[width=\textwidth]{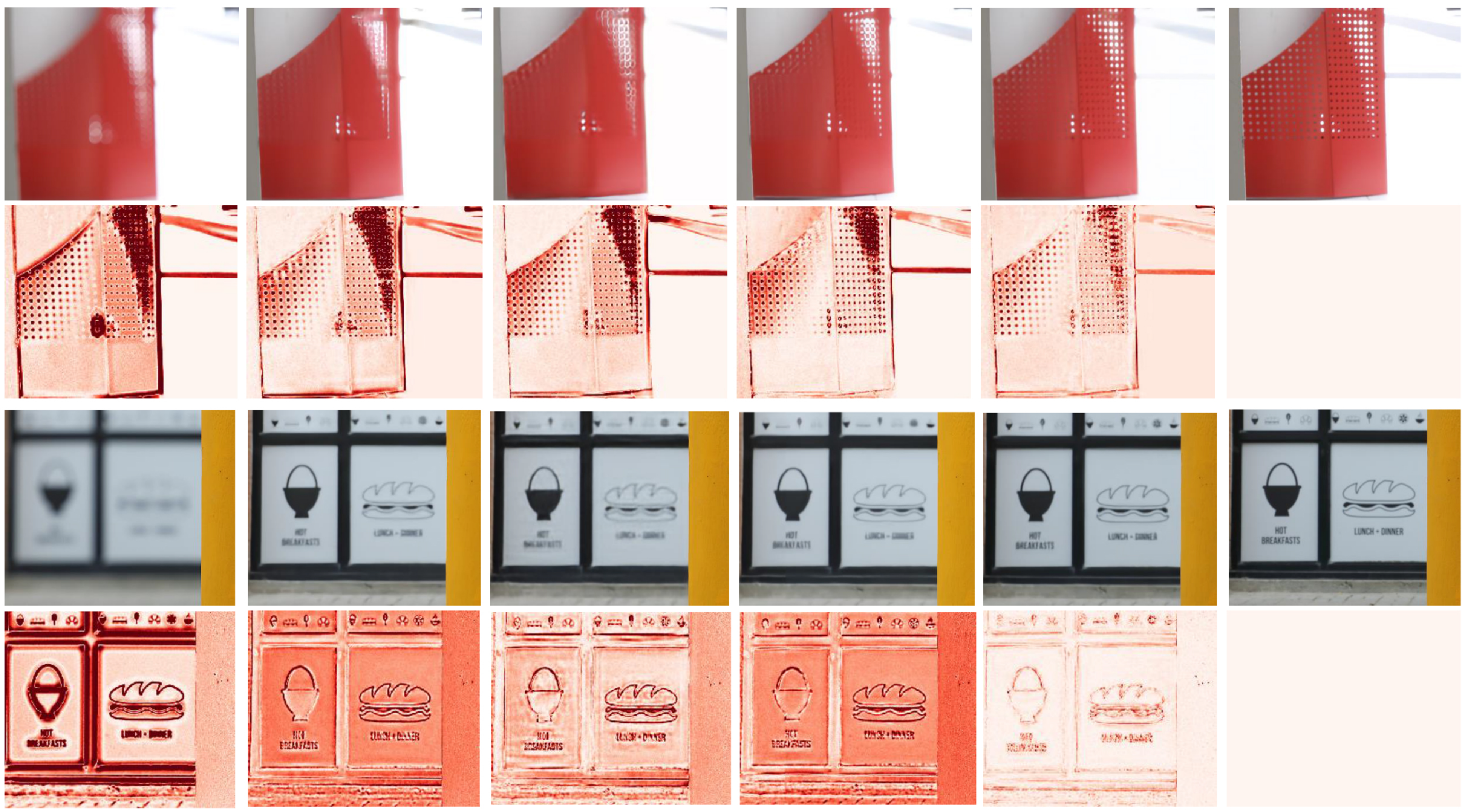}
    \includegraphics[width=\textwidth]{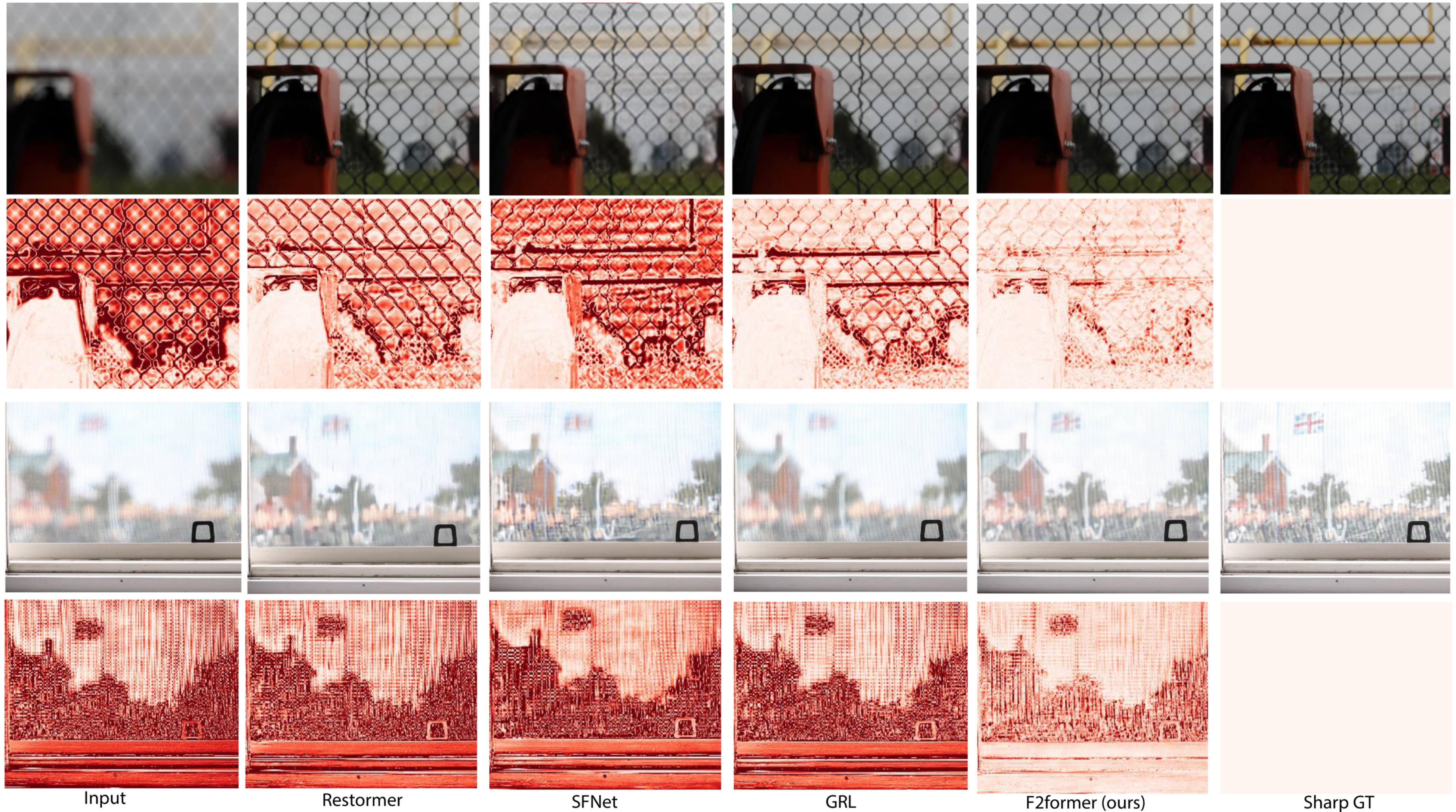}
    \caption{Examples on the DDPD-dual pixel test dataset. The odd rows show blur image, predicted images of different methods, and GT
sharp image. The even rows show the corresponding residual of the blur image / predicted sharp images and GT sharp image. Better visualized at 200\%.}
    \label{fig:ddpd-d}
\end{figure*}

\begin{figure*}
    \centering
    \includegraphics[width=\textwidth]{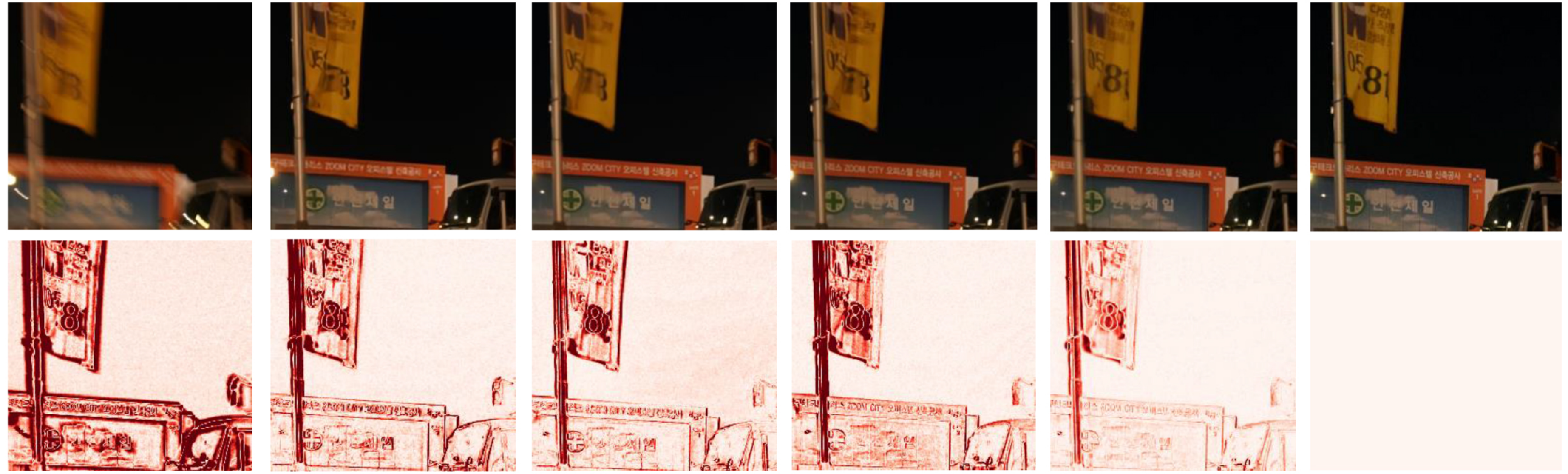}
    \includegraphics[width=\textwidth]{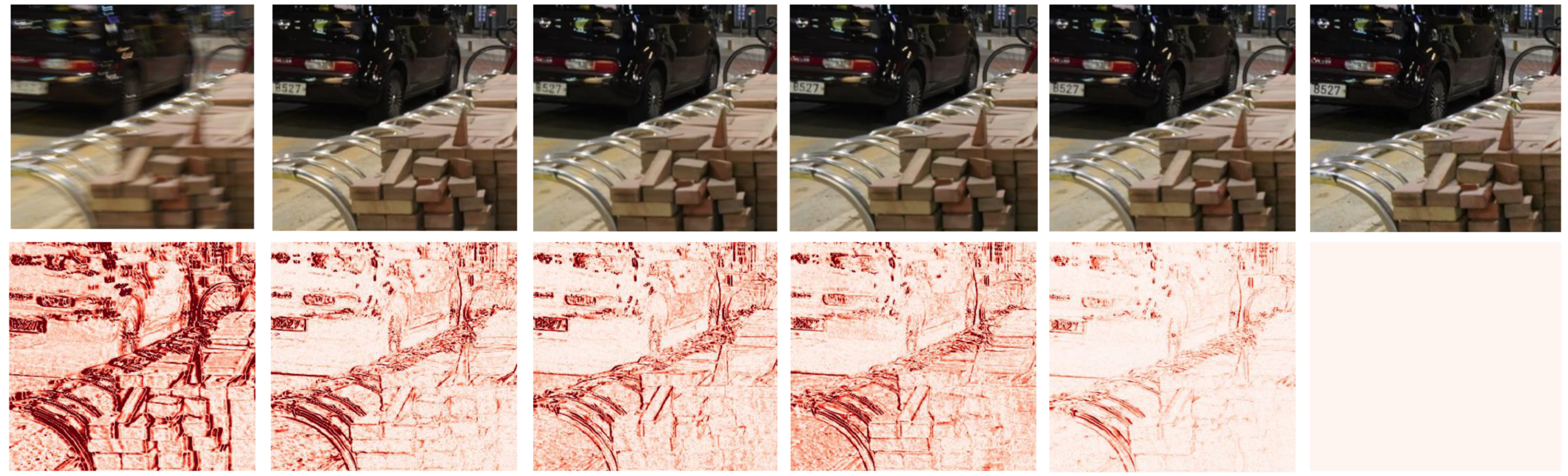}
    \includegraphics[width=\textwidth]{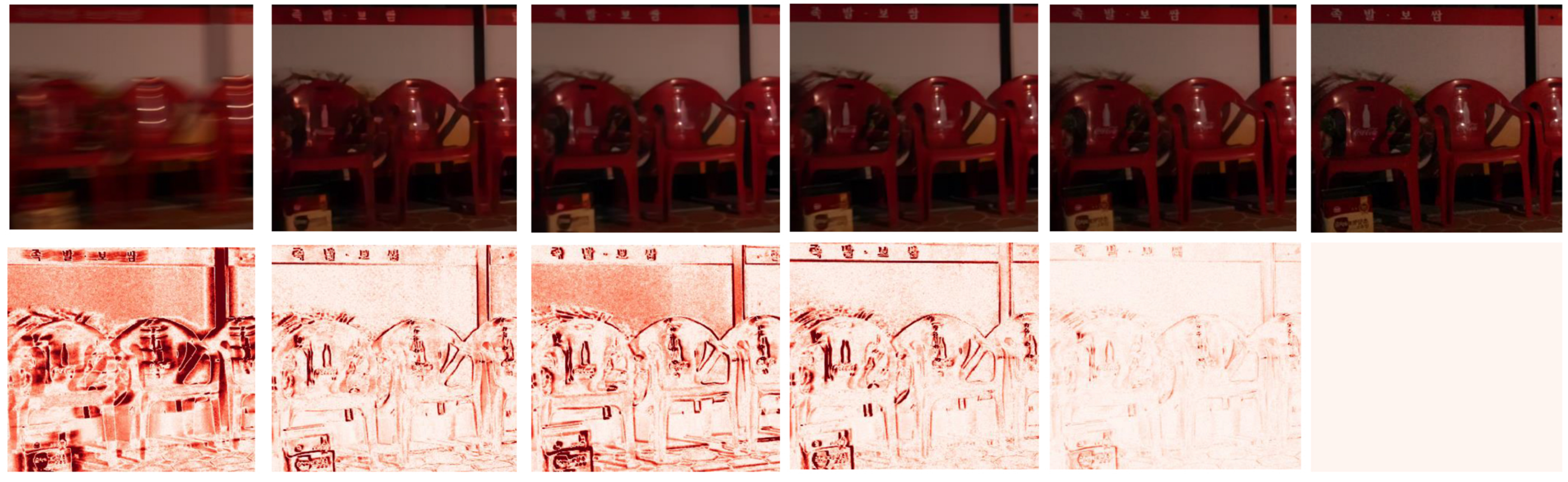}
    \includegraphics[width=\textwidth]{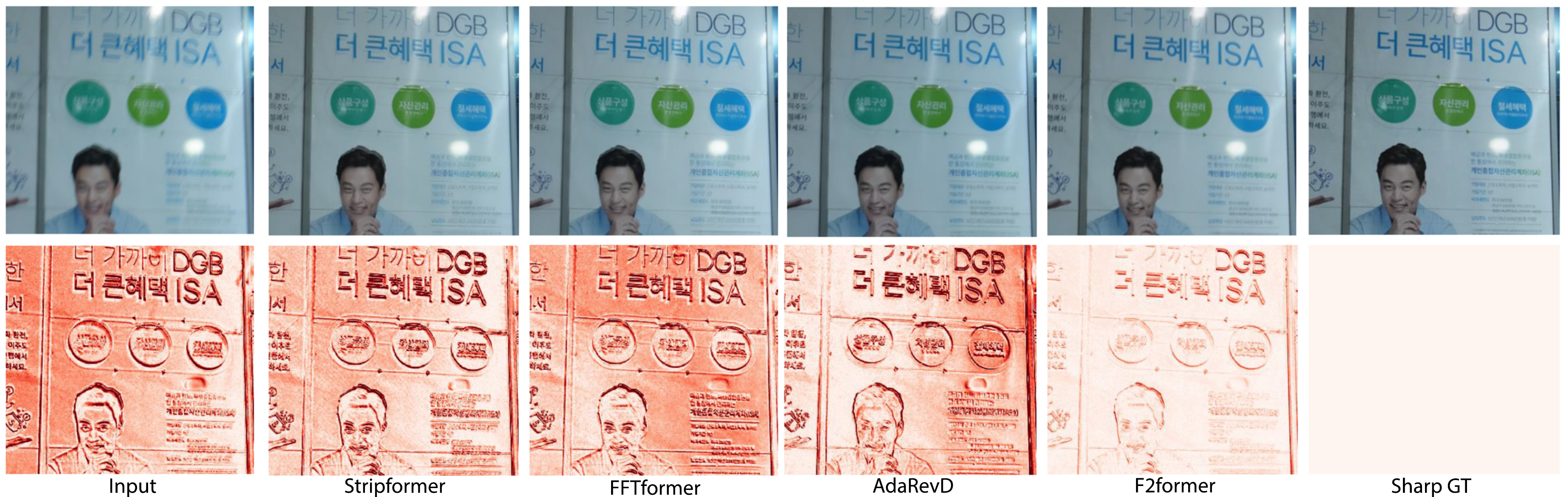}
    \caption{Examples on the RealBlur-J test dataset. The odd rows show blur image, predicted images of different methods, and GT
sharp image. The even rows show the corresponding residual of the blur image / predicted sharp images and GT sharp image. Better visualized at 200\%.}
    \label{fig:realblur}
\end{figure*}

\begin{figure*}
    \centering
    \includegraphics[width=\textwidth]{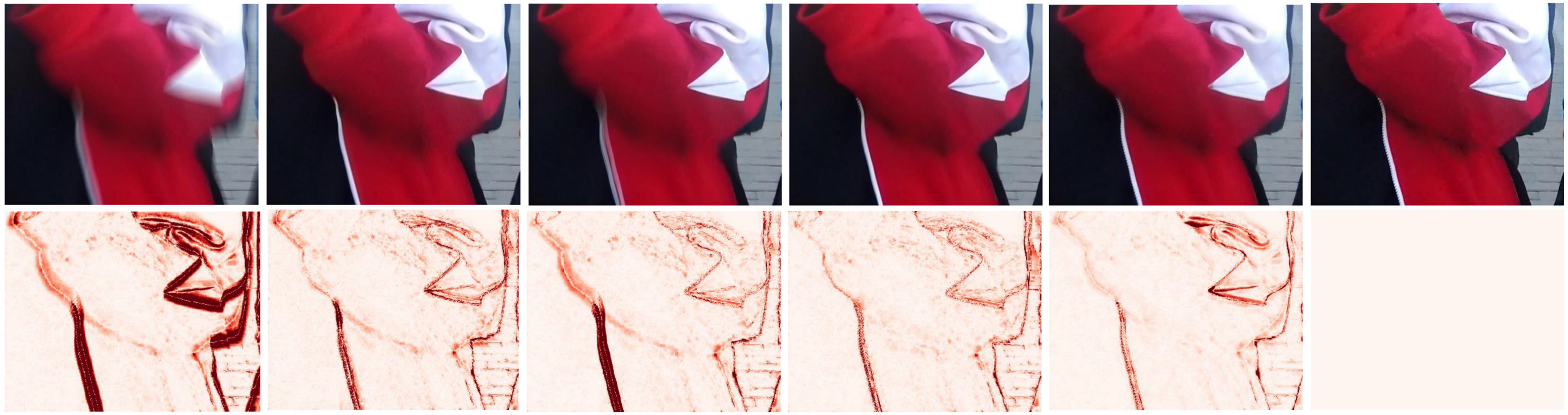}
    \includegraphics[width=\textwidth]{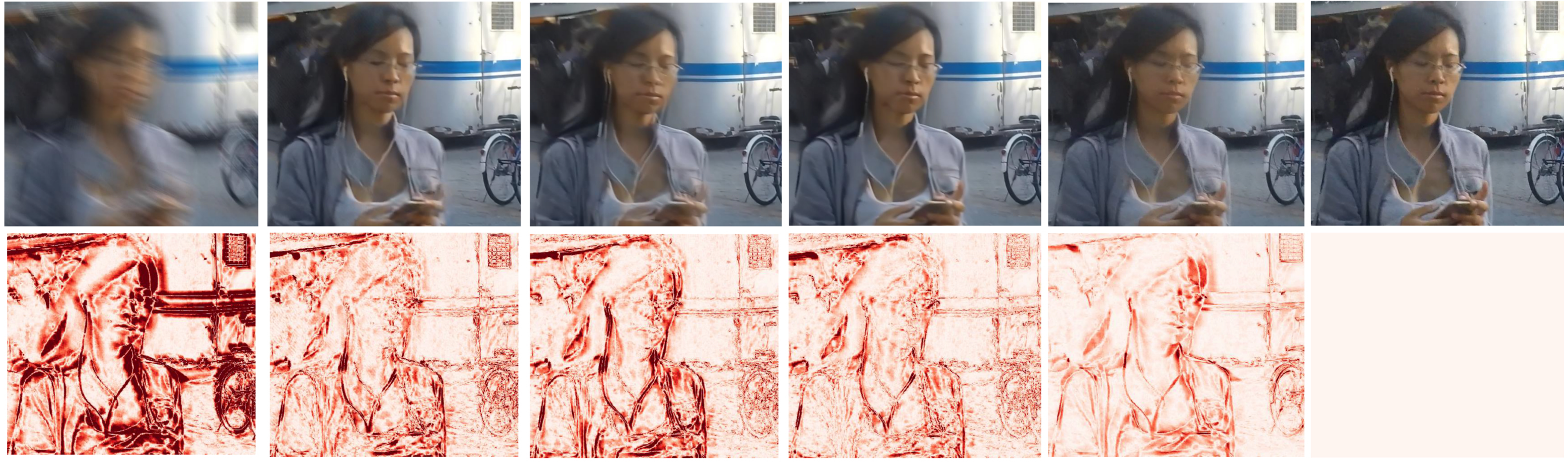}
    \includegraphics[width=\textwidth]{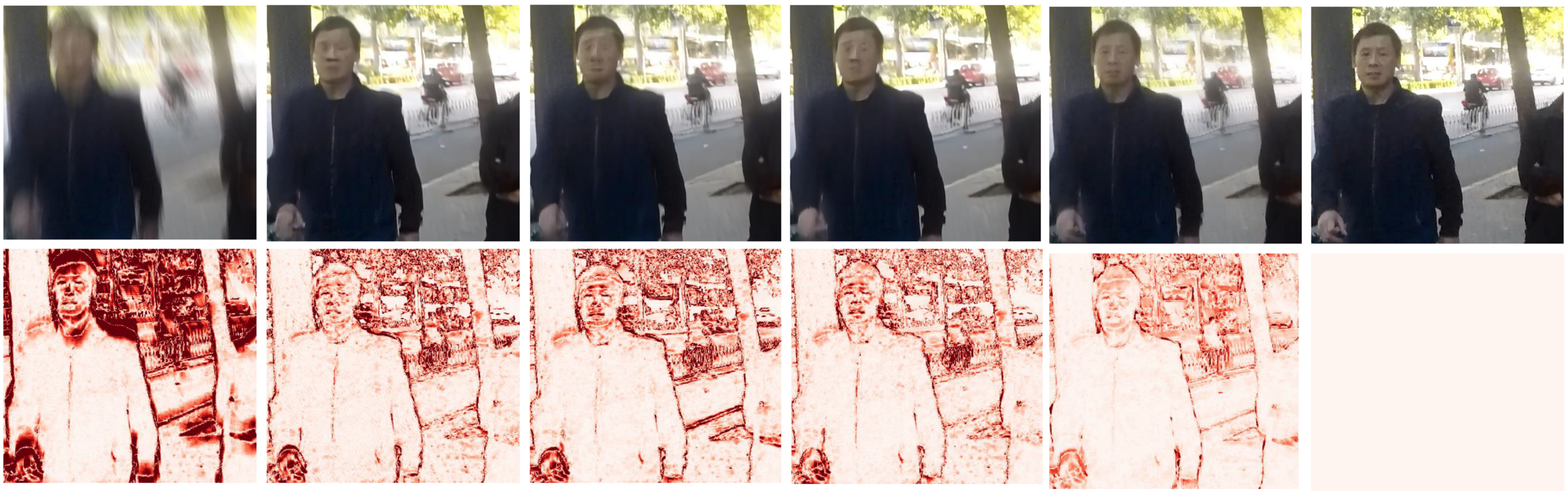}
    \includegraphics[width=\textwidth]{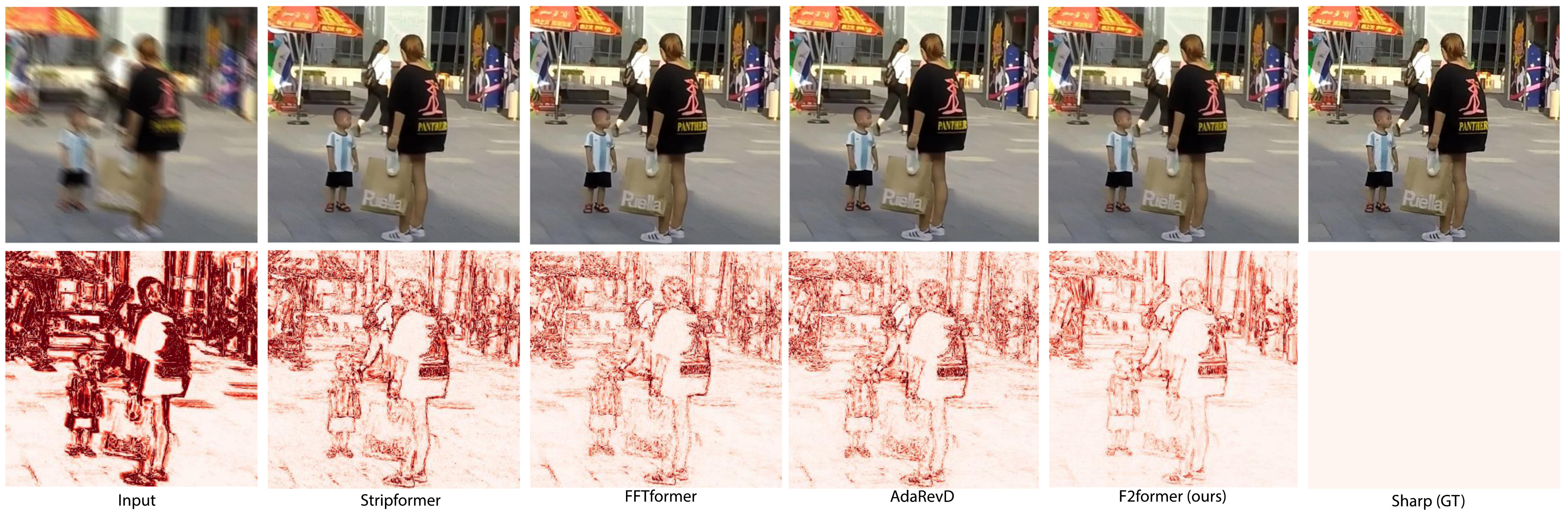}
    \caption{Examples on the HIDE test dataset. The odd rows show blur image, predicted images of different methods, and GT
sharp image. The even rows show the corresponding residual of the blur image / predicted sharp images and GT sharp image. Better visualized at 200\%.}
    \label{fig:hide}
\end{figure*}


\end{document}